\documentclass[journal]{IEEEtran}

\usepackage[pdftex]{graphicx}
\graphicspath{{./images/}}
\DeclareGraphicsExtensions{.pdf,.jpeg,.png}

\usepackage{booktabs}
\usepackage{amsmath}
\usepackage{amssymb}
\usepackage{amsthm}
\usepackage{multirow}
\usepackage{tikz}
\usepackage{xcolor}
\usepackage{fontawesome5} 
\usepackage{algorithm}
\usepackage{algpseudocode}
\usepackage{enumitem}

\usepackage[draft]{changes}
\definechangesauthor[name={Mosab}, color=blue]{MB}
\definechangesauthor[name={Yinghao}, color=orange]{YH}
\definechangesauthor[name={Yinghao}, color=red]{MOVE}

\newcommand{\etal}{\textit{et~al.}}
\newcommand{\NPhard}{\text{NP-hard}}

\usetikzlibrary{arrows.meta, positioning}

\begin{document}

\title{Bilevel Late Acceptance Hill Climbing for the Electric Capacitated Vehicle Routing Problem}

\author{Yinghao~Qin,
        Mosab~Bazargani,
        Edmund~K.~Burke,
        Carlos~A.~Coello~Coello,
        Zhongmin~Song,
        and~Jun~Chen%
\thanks{Y. Qin and J. Chen are with the Centre for Intelligent Transport, QMUL, London, U.K. (e-mail: y.qin@qmul.ac.uk; jun.chen@qmul.ac.uk).}
\thanks{M. Bazargani and E. K. Burke are with the School of Computer Science and Engineering, Bangor University, Bangor, U.K. (e-mail: m.bazargani@bangor.ac.uk; ekb@bangor.ac.uk).}%
\thanks{C. A. Coello Coello is with the Department of Computer Science, Evolutionary Computation Group, CINVESTAV-IPN, Mexico City, Mexico (e-mail: carlos.coellocoello@cinvestav.mx). He is also (as part of a sabbatical leave) with the Basque Center for Applied Mathematics \& Ikerbasque, Spain.}%
\thanks{Z. Song is with the School of Future Technology, Shanghai University, Shanghai, China (e-mail: songzm@shu.edu.cn).}%
\thanks{The first author is supported by the China Scholarship Council (Grant No. 202209110001).}%
\thanks{Manuscript received April 19, 2005; revised August 26, 2015. \textit{(Corresponding author: Jun Chen.)}}}

\markboth{Journal Title}{Qin \MakeLowercase{\etal}: Bilevel Late Acceptance Hill Climbing for the Electric Capacitated Vehicle Routing Problem}

\maketitle

\begin{abstract}
This paper tackles the Electric Capacitated Vehicle Routing Problem (E-CVRP) through a bilevel optimization framework that handles routing and charging decisions separately or jointly depending on the search stage. By analyzing their interaction, we introduce a surrogate objective at the upper level to guide the search and accelerate convergence. A bilevel Late Acceptance Hill Climbing algorithm (b-LAHC) is introduced that operates through three phases: greedy descent, neighborhood exploration, and final solution refinement. b-LAHC operates with fixed parameters, eliminating the need for complex adaptation while remaining lightweight and effective. Extensive experiments on the IEEE WCCI-2020 benchmark show that b-LAHC achieves superior or competitive performance against eight state-of-the-art algorithms. Under a fixed evaluation budget, it attains near-optimal solutions on small-scale instances and sets 9/10 new best-known results on large-scale benchmarks, improving existing records by an average of 1.07\%. Moreover, the strong correlation (though not universal) observed between the surrogate objective and the complete cost justifies the use of the surrogate objective while still necessitating a joint solution of both levels, thereby validating the effectiveness of the proposed bilevel framework and highlighting its potential for efficiently solving large-scale routing problems with a hierarchical structure. 
\end{abstract}


\IEEEpeerreviewmaketitle

\section{Introduction}

\begin{figure}[ht]
	\newcommand{\DepotIcon}{\textcolor{red!60!black}{\Large \faWarehouse}}
	\newcommand{\StoreIcon}{\textcolor{gray!90!black}{\large\faStore}}
	\newcommand{\ChargingStationIcon}{\textcolor{green!60!black}{\large\faChargingStation}}
	\centering
    \scalebox{0.85}{ 
	\begin{tikzpicture}[>=latex]
		
		\node (depot) at (3.5,3.5) {\DepotIcon};
		
		\node (shop1) at (0,3.5) {\StoreIcon};
		\node (shop2) at (1.5,4.5) {\StoreIcon};
		\node (shop3) at (1.5,2.5) {\StoreIcon};
		\node (shop4) at (3,5.5) {\StoreIcon};
		\node (shop5) at (5,4.5) {\StoreIcon};
		\node (shop6) at (5.5,7) {\StoreIcon};
		\node (shop7) at (7,5.5) {\StoreIcon};
		\node (shop8) at (7,4) {\StoreIcon};
		\node (shop9) at (4,2) {\StoreIcon};
		\node (shop10) at (6,1.5) {\StoreIcon};
		
		\node (charger1) at (8,1.5) {\ChargingStationIcon};
		\node (charger2) at (2.5,1.5) {\ChargingStationIcon};
		\node (charger3) at (6,3) {\ChargingStationIcon};
		\node (charger4) at (5,5.7) {\ChargingStationIcon};
		
		\draw[->, thick] (depot) -- (shop2);
		\draw[->, thick] (shop2) -- (shop1);
		\draw[->, thick] (shop1) -- (shop3);
		\draw[->, thick] (shop3) -- (depot);
		
		\draw[->, dashed, thick] (depot) -- (shop5);
		\draw[->, dashed, thick] (shop5) -- (charger4);
		\draw[->, dashed, thick] (charger4) -- (shop7);
		\draw[->, dashed, thick] (shop7) -- (shop6);
		\draw[->, dashed, thick] (shop6) -- (charger4);
		\draw[->, dashed, thick] (charger4) -- (shop4);
		\draw[->, dashed, thick] (shop4) -- (depot);
		
		\draw[->, dotted, thick] (depot) -- (shop9);
		\draw[->, dotted, thick] (shop9) -- (shop10);
		\draw[->, dotted, thick] (shop10) -- (charger1);
		\draw[->, dotted, thick] (charger1) -- (charger3);
		\draw[->, dotted, thick] (charger3) -- (shop8);
		\draw[->, dotted, thick] (shop8) -- (depot);
		
	\end{tikzpicture}
    }
    \caption{Illustrative solution to the E-CVRP, featuring one depot, four charging stations, and ten customer locations.}
    \label{fig:E-CVRP_example}
\end{figure}
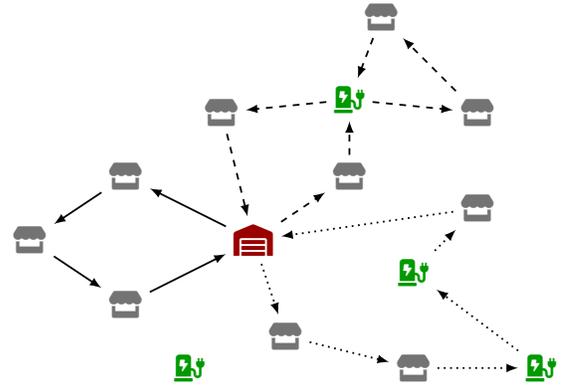


\IEEEPARstart{A}{s} the global shift toward sustainable transportation accelerates, electric vehicles (EVs) have become central to decarbonizing road-based logistics. In this context, the Electric Vehicle Routing Problem (EVRP) has emerged as a key research frontier. Its core variant, the Electric Capacitated Vehicle Routing Problem (E-CVRP)~\cite{mavrovouniotis2020techreport}, forms the foundation of EV logistics. Extensions incorporate features such as non-linear charging, heterogeneous fleets, and time-window constraints. Methodological advances in solving E-CVRP are therefore crucial, underpinning applications from urban last-mile delivery to long-haul freight.

Given a fleet of homogeneous EVs, the objective of E-CVRP is to determine the most efficient route for each vehicle that respects battery charge and vehicle cargo capacity constraints. Each vehicle must begin and end at a depot while serving a set of customers. Fig.~\ref{fig:E-CVRP_example} illustrates a schematic example of a feasible E-CVRP solution. In this example, the first route (solid lines) requires no charging stops. The second route (dashed lines) visits the same charging station twice. The third route (dotted lines) makes consecutive stops at two distinct charging stations before reaching the next customer, representing an extreme case of ultra-long-distance delivery.

The E-CVRP can be viewed as a combination of the classical Capacitated Vehicle Routing Problem (CVRP)~\cite{toth2014vehicle} and the Green Vehicle Routing Problem (GVRP)~\cite{erdougan2012green}. The former considers cargo capacity constraints, whereas the latter incorporates limited driving range under the assumption of unlimited cargo capacity. The E-CVRP simultaneously accounts for both vehicle capacity and driving range constraints. It is also a member of the family of Vehicle Routing Problems with Intermediate Stops (VRPIS)~\cite{schiffer2019vehicle}, where optional en-route stops are introduced to support service operations. VRPIS generally arise in three application contexts: replenishment and disposal operations, rest or idling periods, and refueling. The E-CVRP belongs to the refueling category.

Current methodologies for solving the E‑CVRP face two major challenges. First, at the modeling level, existing formulations either produce intractably large search spaces~\cite{mavrovouniotis2020techreport} or rely on aggressive pruning strategies~\cite{froger2019improved} that risk discarding high‑quality solutions. They often fail to scale effectively to large‑scale instances. Second, from an algorithmic perspective, most existing approaches rely on highly customized, problem‑specific heuristics, often embedded within vague conceptual frameworks~\cite{jia2021bilevel, jia2022confidence, rodriguez2024new, chen2024efficient}. While effective, such heuristics are difficult to reproduce, tune, and adapt. These limitations motivate a modeling approach that reduces computational complexity, together with a transparent, general-purpose solution approach capable of scaling to large instances.

At the root, these difficulties stem from domain characteristics specific to electric fleets. Compared to the classical CVRP, the main additional challenge in E-CVRP arises from the limited cruising range imposed by EV battery capacity. Unlike internal combustion engine vehicles, which can be refueled within minutes, EVs require significantly longer charging times, making detours for recharging costly in terms of operational time and scheduling. In real-world scenarios, EVs may even need to interrupt their routes for multiple recharges within a single trip. From the charging infrastructure perspective, the limited number of stations implies that each of them may be visited once, multiple times, or not at all, possibly by the same or different vehicles. These interdependencies between vehicle routing, battery management, and charging infrastructure create unique modeling challenges and significantly increase the computational complexity in solving E-CVRP at a large scale.

Most existing formulations model the E-CVRP as a \emph{single-level} mixed-integer linear programming (MILP). A representative example is provided by Mavrovouniotis \etal~\cite{mavrovouniotis2020techreport}, where the problem is defined on a \emph{simple graph} and binary decision variables are associated with each arc to indicate whether it is traversed by an EV. In order to accommodate multiple visits to charging stations within a route, charging nodes must be replicated. In some models, the number of replications is bounded by twice the number of customers~\cite{mavrovouniotis2020benchmark}, corresponding to a worst-case scenario in which a vehicle visits a charging station both before and after serving each customer. Although mathematically rigorous, such a node replication approach dramatically enlarges the search space and makes the problem more difficult to solve. To overcome these drawbacks, Froger~\etal~\cite{froger2019improved} propose an alternative formulation based on pre-enumerating feasible charging station paths (CSPs) between each pair of non-charging nodes (customers or depots). The E-CVRP is then represented as a \emph{multigraph}, where multiple arcs between a pair of non-charging nodes correspond to different feasible CSPs. The model then selects appropriate CSPs to connect non-charging nodes and thereby construct complete EV routes. However, the number of feasible CSPs grows rapidly with instance size. Even with strong dominance pruning, hundreds of paths are generated for 10-customer instances and over a thousand for 20-customer instances on average~\cite{froger2019improved}. Consequently, the resulting MILP can be solved using a commercial solver only for small-scale instances, while most 20-customer cases already become computationally intractable.

In light of these limitations of single-level formulations, we adopt a bilevel modeling perspective for the E-CVRP. The proposed formulation decomposes the problem into two interdependent components: routing and charging. We analyze their interactions and leverage these insights to guide algorithm design, focusing on the most promising regions of the search space. At the upper level, a CVRP is solved, where customers are partitioned into feasible subsets and the visiting sequence is determined for each subset. At the lower level, a fixed-route vehicle charging problem (FRVCP)~\cite{montoya2017electric} optimizes charging station insertions along the given path from the upper level. This bilevel structure substantially reduces the computational complexity and provides a theoretical framework for integrated routing and charging decision making.

We further develop a bilevel Late Acceptance Hill Climbing (b-LAHC) algorithm for the E-CVRP. LAHC is a single-point metaheuristic~\cite{burke2017late} that has shown success in various scheduling and combinatorial optimization problems~\cite{fonseca2016late,bolaji2018late}. Its core mechanism, the ``late acceptance'' rule~\cite{burke2017late}, is simple yet effective: a candidate solution is accepted if it improves upon the current solution or outperforms the one encountered several iterations earlier. This mechanism introduces a well-balanced trade-off between exploration and exploitation. 

The main contributions of this paper are the following:
\begin{enumerate}[leftmargin=1em, itemsep=0ex]
    \item A \emph{bilevel optimization model} is explicitly adopted for the E-CVRP for the first time. The model captures both routing and charging decisions without duplicating nodes for charging stations. The solution and objective spaces are further analyzed to guide efficient search strategies. Notably, the model enables \emph{partial evaluation} of solutions without solving the lower level, which serves as a surrogate for full evaluation and accelerates the search.
    \item A bilevel metaheuristic, b-LAHC, is proposed based on late acceptance hill climbing. It features three phases: (i) \emph{initialization}, where a greedy descent rapidly drives the upper-level routing solution to a local optimum; (ii) \emph{exploration}, a neighborhood search guided by the late acceptance mechanism at the upper level, with efficient charging decisions conditionally triggered at the lower level; and (iii) \emph{refinement}, where exhaustive charging optimization is applied to the best-found solution at termination. This three-phase design balances computational efficiency and solution quality, while inheriting the convergence properties of the original LAHC and adapting them to the bilevel structure. 
    \item An extensive study is conducted on the IEEE WCCI 2020 E-CVRP benchmark~\cite{mavrovouniotis2020techreport}, comparing b-LAHC with seven state-of-the-art algorithms across 17 instances under two termination criteria: a fixed number of evaluations and a fixed runtime. Overall, b-LAHC achieves \emph{10 new best-known solutions} (BKS), matches 2 and remains competitive on the rest. Its strength is particularly pronounced on large-scale instances, improving the current BKS by an average of \emph{1.07\%} across 10 cases, and reducing solution standard deviation by \emph{46.2\%}, relative to the second-best method.
\end{enumerate}

The remainder of this paper is organized as follows. Section~\ref{sec:related-work} reviews the literature on bilevel optimization in routing problems and summarizes the current approaches for EVRPs. Section~\ref{sec:mathematical-modeling} presents the proposed bilevel mathematical model of the E-CVRP. Section~\ref{sec:algorithm} introduces the b-LAHC algorithm. Section~\ref{sec:experiments} reports our computational experiments and analyzes the results in detail. Section~\ref{sec:conclusion} concludes the paper and outlines potential directions for future research.

\section{Related Work}
\label{sec:related-work}

\begin{table*}[!t]
\centering
\caption{Summary of notation used in the bilevel E-CVRP model.}
\label{tab:notation}
\small
\renewcommand{\arraystretch}{0.8}
\begin{tabular}{p{0.09\textwidth} p{0.22\textwidth} p{0.62\textwidth}}
\toprule
\textbf{} & \textbf{Symbol} & \textbf{Description} \\
\midrule
\multirow{10}{*}{General}
  & $G = (V, A)$ & Complete graph representing the E-CVRP instance. \\
  & $V = \{d\} \cup \mathcal{V}_c \cup \mathcal{V}_f$ & Node set consisting of the depot $d$, customer set $\mathcal{V}_c$, and charging station set $\mathcal{V}_f$. \\
  & $A = \{(i, j) \mid i \ne j,\ i, j \in V\}$ & Set of arcs representing all possible connections between distinct nodes. \\
  & $\delta_i$ & Demand of customer $i \in \mathcal{V}_c$. ($\delta_i > 0$) \\
  & $d_{ij}$ & Distance between nodes $i$ and $j$, for all $(i, j) \in A$. ($d_{ij} \ge 0$) \\
  & $h$ & Battery consumption rate per unit distance. ($h > 0$) \\
  & $Q_c$ & Maximum cargo capacity of the EV. ($Q_c > 0$) \\
  & $Q_b$ & Maximum battery capacity of the EV. ($Q_b > 0$) \\
  & $\sigma_i$ & State-of-charge (SoC) of the EV upon departure from node $i \in V$. ($\sigma_i \ge 0$) \\
  & $M$ & The available number of identical EVs. ($M > 0$) \\
\midrule
\multirow{8}{*}{Upper Level}
  & $R_v = \{n_0^v,\dots, n_{L_v+1}^v\}$ & Ordered sequence of nodes visited by vehicle $v$, starting and ending at the depot $d$ ($n_0^v = n_{L_v+1}^v = d$), where $L_v$ denotes the number of customers served. \\
  & $\bar{R}_v = R_v \oplus S_v$ & Complete route of vehicle $v$ obtained by integrating charging decisions $S_v$ into the customer route $R_v$. \\
  & $x = \{R_1, \dots, R_M\}$ & Upper-level decision variable representing a set of $M$ customer-serving routes, where some routes $R_v$ may be empty, indicating that vehicle $v$ is not used. \\
  & $\mathcal{X}$ & Feasible set of all upper-level routing solutions $x$. \\
  & $F(x, y^*(x))$ & Total objective value of a solution with routing plan $x$ and its optimal charging $y^*(x)$. \\
\midrule
\multirow{11}{*}{Lower Level}
  & $S_v = \{s_0^v, \dots, s_{L_v}^v\}$ & The charging decision along route $R_v$, where each $s_\ell^v$ represents all possible charging station visit configurations between the paired nodes $(n_\ell^v, n_{\ell+1}^v)$. 
  \begin{itemize}[leftmargin=1.5em]
    \item If at most one charging station is allowed: $s_\ell^v \in \mathcal{V}_f \cup \{\texttt{NIL}\}$.
    \item If up to two charging stations are allowed: $s_\ell^v \in \mathcal{V}_f \cup \{(u,\!w) \mid u,\!w \in \mathcal{V}_f,\! u \ne w\} \cup \{\texttt{NIL}\}$, where $(u, w)$ denotes an ordered pair of distinct charging stations inserted between two nodes in a route.
  \end{itemize}
  Here, \texttt{NIL} indicates that no charging station is inserted. \\
  & $y(x) = \{S_1, \dots, S_M\}$ & Lower-level decision variable representing the charging decisions for all $M$ routes in the routing plan $x$. \\
  & $\mathcal{Y}(x)$ & Feasible set of all lower-level charging solutions $y$ given a routing plan $x$. \\
  & $f(x, y)$ & Extra travel cost caused by charging detours for routing plan $x$ under decision $y$. \\
\bottomrule
\end{tabular}
\end{table*}

\subsection{Solution Approaches for EVRPs}
Depending on the scale and realism of modern EVRPs, solution approaches broadly fall into two families: exact and heuristic. Each offers complementary strengths and trade-offs.

Exact methods are highly effective for small-scale instances but scale poorly due to exponential complexity. Most rely on MILP formulations and commercial solvers or specialized branch-and-bound frameworks. Zuo~\etal~\cite{zuo2019new} employed secant-based piecewise linearization to approximate the concave nonlinear charging, solving instances with up to 25 customers and 4 stations using CPLEX. Kancharla and Ramadurai~\cite{kancharla2020electric} extended this with nonlinear charging and load-dependent discharging, solving instances with up to 30 customers and 4 stations with Gurobi. Tahami~\etal~\cite{tahami2020exact} applied branch-and-cut to E-CVRP instances with up to 30 customers and 21 charging stations. More recently, Lam~\etal~\cite{lam2022evrptw} proposed a branch-and-cut-and-price for an Electric Vehicle Routing Problem with Time Windows (EVRPTW) that incorporates piecewise-linear recharging and capacitated charging stations, tested on instances with up to 100 customers. Nafstad~\etal~\cite{nafstad2025bpc} developed a branch-price-and-cut for EVRPTW with heterogeneous charging technologies and nonlinear functions, optimally solving instances with up to 100 customers and 21 stations within one hour. These studies adopt highly detailed modeling; when combined with exact algorithms, such models further compound the computational burden, making it particularly difficult to scale to larger, real-world instances.

In contrast to exact algorithms, heuristic methods dominate for medium- and large-scale EVRPs, where exact solvers become impractical. Their efficiency and robustness lead to high-quality approximate solutions within practical time, a key consideration for many real-world applications.  

A wide range of heuristic and metaheuristic frameworks have been developed for the EVRP and its variants, spanning both population-based and single-point algorithms. In the population-based category, Caillard and Ben Chabane~\cite{caillard2024evolutionary} introduced a hybrid approach that combines genetic mechanisms with ACO for the dynamic EVRP with time windows. Yang~\etal~\cite{yang2015electric} developed a partheno-genetic algorithm to minimize total distribution costs under time-of-use electricity pricing, considering both fast and regular charging options. Zhen~\etal~\cite{zhen2020hybrid} proposed a Particle Swarm Optimization algorithm for a variant of EVRP with different driving modes. In addition, several single-point metaheuristics have been proposed to EVRPs. Felipe~\etal~\cite{felipe2014heuristic} used SA for a Green VRP variant that permits partial recharging and accommodates heterogeneous charging technologies. Montoya~\etal~\cite{montoya2017electric} developed an Iterated Local Search algorithm to optimize nonlinear charging decisions along predetermined vehicle routes. Goeke~\cite{goeke2019granular} proposed a granular Tabu Search for the EV Pickup and Delivery Problem with Time Windows.

Furthermore, a number of hybrid approaches have also been developed for EVRPs. Seyfi~\etal.~\cite{seyfi2022multi} studied the Multi-Mode Hybrid EVRP using a matheuristic that combines Variable Neighborhood Search and mathematical programming. Keskin and Çatay~\cite{keskin2018matheuristic} addressed an EVRP with time windows, multiple charging speeds and partial recharges, solving it with an Adaptive Large Neighborhood Search (ALNS) and a MILP metaheuristic. Nolz~\etal~\cite{nolz2022consistent} introduced the consistent EVRP with backhauls and charging-slot management, and developed a hybrid framework that integrates ALNS with constraint programming for charging slots and quadratic programming for trip scheduling. Overall, existing approaches rely either on computationally intensive formulations or heavily customized heuristics, limiting scalability and transparency.

\subsection{Bilevel Optimization in Vehicle Routing Problems}
Many real-world routing problems exhibit hierarchical decision structures in which decisions at one level depend on and influence those at another. While such problems can be formulated as single-level models, doing so often results in inefficiencies and reduced interpretability. Bilevel optimization offers an alternative modeling paradigm for hierarchical routing problems. It reflects a divide-and-conquer idea, decomposing a complex problem into interdependent subproblems that are solved in a coordinated manner. Furthermore, when the resulting subproblems have been well studied in the literature, existing solution techniques can be directly leveraged rather than developing entirely new algorithms from scratch. This allows methodological efforts to focus more on how to decompose the problem itself and how to exploit the interaction and coordination between the two levels.

Nearly two decades ago, Marinakis~\etal~\cite{marinakis2007new} proposed one of the earliest bilevel formulations for the CVRP, combining a Set Partitioning Problem at the upper level and a Traveling Salesman Problem at the lower level, solved by a Genetic Algorithm (GA). Tu~\etal~\cite{tu2014bi} later proposed a bilevel Simulated Annealing (SA) for the large-scale Multi-Depot Vehicle Routing Problem, where the upper level assigns customers to depots via Voronoi cells (``cluster-first, route-second''~\cite{gillett1974heuristic}) and the lower level refines routes through reassignment and local search. Zhou~\etal~\cite{zhou2023bilevel} developed a bilevel memetic algorithm for the Soft-Clustered VRP, with cluster assignments at the upper level and customer routing at the lower level, integrating group matching-based crossover, bilevel neighborhood search, and tabu-based reconstruction. Jia~\etal~\cite{jia2021bilevel,jia2022confidence} introduced a bilevel ant colony optimization (ACO) approach for the E-CVRP, combining an ``order-first, split-second'' method~\cite{prins2014order} with local search at the upper level and feasible recharging schedules at the lower level. Qin and Chen~\cite{qin2024confidence} further proposed a memetic algorithm with adaptive selection, achieving state-of-the-art performance with 6 new best-known solutions on the 17-instance IEEE WCCI-2020 benchmark~\cite{mavrovouniotis2020techreport}. Although these studies decompose the E-CVRP into routing and charging subproblems, the interaction between the two levels has not been systematically analyzed.

\section{Mathematical Modeling}
\label{sec:mathematical-modeling}

The E-CVRP is defined on a directed complete graph \(G=(V,A)\) with symmetric distances \(d_{ij}=d_{ji}\), where \(V=\{d\}\cup \mathcal{V}_c \cup \mathcal{V}_f\) and \(A=\{(i,j)\mid i\neq j,\ i,j\in V\}\). Here, \(d\) denotes the depot, \(\mathcal{V}_c\) the set of customers, and \(\mathcal{V}_f\) the set of charging stations. Each customer \(i\in\mathcal{V}_c\) has demand \(\delta_i>0\). Traveling from \(i\) to \(j\) consumes \(h\,d_{ij}\) units of energy, where \(h\) is the per-distance energy rate. Each EV is characterized by cargo capacity \(Q_c\) and battery capacity \(Q_b\). We let \(\sigma_i\) denote the battery level upon departure from node \(i\) and \(M\) the number of available identical EVs.

\subsection{Bilevel Modeling}\label{subsec:bilvel-modeling}

The E-CVRP is formulated using a bilevel model involving two hierarchical decision-making levels. The upper-level decision maker, referred to as the \emph{leader}, controls customer-serving routes. The lower-level decision maker, the \emph{follower}, determines feasible charging strategies following the routes specified by the \emph{leader}. The complete bilevel optimization model is formulated as follows, with all relevant notations defined in Table~\ref{tab:notation}. 

\vspace{1em} 
\noindent
\textbf{Upper-level (routing decision):}
\begin{equation}
\min_{x \in \mathcal{X}} \; F(x, y^*(x)),
\quad \text{where } F(x, y^*(x)) = \sum_{v=1}^M \text{cost}(\bar{R}_v)
\label{eq:upper-obj}
\end{equation}

\noindent
\textbf{Subject to:}
\begin{align}
& \bigcup_{v=1}^M \left(R_v \setminus \{d\}\right) = \mathcal{V}_c 
   \tag{2} \label{constraints:customer-service} \\
& \left(R_u \setminus \{d\}\right) \cap \left(R_v \setminus \{d\}\right) = \emptyset,
   \quad \forall\, u, v \in \{1,\dots,M\},\ u \ne v
   \tag{3} \label{constraints:cap} \\
& \sum_{i \in R_v} \delta_i \le Q_c,
   \quad \forall\, v = 1,\dots,M
   \tag{4} \label{constraints:capacity}
\end{align}

\noindent
\textbf{Lower-level (charging decisions for a given \(x\)):}
\begin{equation}
y^*(x) = \arg\min_{y \in \mathcal{Y}(x)} f(x, y),
\quad \text{where } f(x, y) = \sum_{v=1}^M \text{cost}(S_v) \tag{5}
\label{eq:lower-obj}
\end{equation}

\noindent
\textbf{Subject to:}
\begin{align}
& \sigma_i = Q_b, \quad \forall i \in V_f \cup \{d\} \tag{6} \label{constraints:station-visit}\\
& \sigma_j = \sigma_i - h \cdot d_{ij}, \quad \forall (i, j) \in \bar{R}_v \tag{7} \label{constraints:battery-transition} \\
& 0 \le \sigma_i \le Q_b, \quad \forall i \in \bar{R}_v \tag{8} \label{constraints:battery-feasibility}
\end{align}

The upper-level objective~\eqref{eq:upper-obj} seeks a routing plan \( x \) that minimizes the distance traveled by all vehicles, under the assumption that the \emph{follower} always computes the best feasible charging strategy \( y^*(x) \) in response. Constraints~\eqref{constraints:customer-service}-\eqref{constraints:capacity} ensure that all customers are exclusively assigned to EVs and that the demand on each route does not exceed the vehicle's capacity.

Given a routing plan $x$, the \emph{follower} aims to determine the charging strategy \( y^*(x) \) that minimizes the additional distance traveled when making a detour to charging stations~\eqref{eq:lower-obj}. Constraints~\eqref{constraints:station-visit}-\eqref{constraints:battery-feasibility} ensure battery feasibility: EV departs from the depot or charging station with a full battery, maintains sufficient charge between consecutive nodes, and keeps the battery level within valid bounds throughout the route.

The resulting optimal response $y^*(x)$ is passed back to the \emph{leader}, who updates the upper-level objective accordingly. This interaction continues iteratively until termination, yielding an approximate solution pair $(x^*, y^*)$.

Notably, to cover all feasible configurations in the \emph{follower}'s charging decisions, it is necessary to bound the number of charging stations inserted between any two successive nodes. Thus, we introduce the following assumption: \emph{For any two successive nodes in a route, the EV requires no more than two charging station visits to traverse the segment.}

This assumption aligns with practical applications, as EVs are generally used for short- to medium-range travel, while long-distance routes that may require more than two charging stops are typically served by fuel-based vehicles.


\subsection{Solution Space Analysis}\label{subsec:solution-space-analysis}

\begin{figure}[!t]
  \centering
  \includegraphics[width=0.85\linewidth]{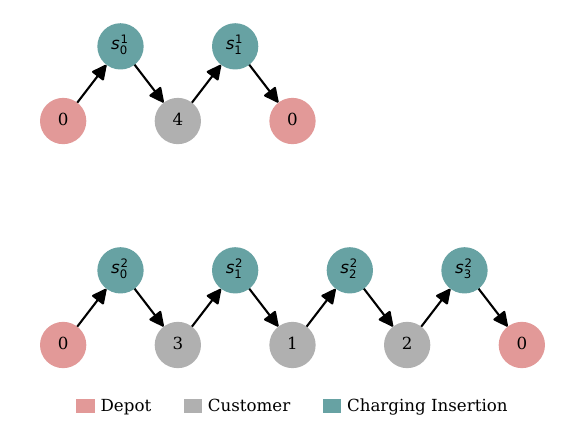}
  \caption{An illustrative example of a solution representation for the E-CVRP.}
  \label{fig:solution-representation}
\end{figure}

Fig.~\ref{fig:solution-representation} illustrates a sample solution for the E-CVRP instance with depot~0, customers~1-4, and charging stations~5 and~6. The upper-level solution $\mathbf{x}$ consists of two routes: $R_1$: $0 \rightarrow 4 \rightarrow 0$ and $R_2$: $0 \rightarrow 3 \rightarrow 1 \rightarrow 2 \rightarrow 0$. In the corresponding lower-level decision $y(\mathbf{x})$, each route $R_v$ is associated with a charging decision sequence $S_v$, which together form the complete route $\bar{R}_v$. Specifically, $\bar{R}_1$ is given by $0 \rightarrow s^1_0 \rightarrow 4 \rightarrow s^1_1 \rightarrow 0$, and $\bar{R}_2$ is given by $0 \rightarrow s^2_0 \rightarrow 3 \rightarrow s^2_1 \rightarrow 1 \rightarrow s^2_2 \rightarrow 2 \rightarrow s^2_3 \rightarrow 0$. Each $s^v_\ell$ denotes a charging decision after leaving node~$\ell$ on route~$R_v$, where $s^v_\ell$ can be selected from the set $\{\texttt{NIL}, 5, 6, (5,6), (6,5)\}$.

Upper-Level Space $\mathcal{X}$: The upper level determines a set of routes $x = \{R_1, \dots, R_M\}$, where each route $R_v$ is an ordered sequence of customer nodes. This decision involves both the partitioning of the customer set and the permutation of nodes within each route. The number of possible partitions increases super-exponentially (Bell number), and the number of permutations per route grows factorially. Therefore, the size of $\mathcal{X}$ can be approximated as:
$$
|\mathcal{X}| \approx \text{Partition}(V_c) \times \prod_{v=1}^{M} L_v!
$$
where \(L_v\) is the number of customers served by vehicle \(v\).

Lower-Level Space $\mathcal{Y}(x)$: Given a routing plan $x$, the lower level determines where to insert charging stations along each route to ensure battery feasibility. For a route $R_v$ with $L_v+1$ insertion points (between successive nodes), and assuming that up to two stations may be inserted in each gap, the number of insertion options per insertion point is:
$$
1 + |\mathcal{V}_f| + |\mathcal{V}_f| (|\mathcal{V}_f| - 1)
$$
Each term of the above corresponds to no insertion, one station, or a pair of distinct stations. Therefore, the total number of feasible insertions for route $R_v$ is:
$$
|\mathcal{Y}_v| = \left(1 + |\mathcal{V}_f| + |\mathcal{V}_f|(|\mathcal{V}_f|-1)\right)^{L_v+1}
$$
and the full lower-level space is:
$$
|\mathcal{Y}(x)| = \prod_{v=1}^{M} |\mathcal{Y}_v|
$$

The total solution space of the bilevel E-CVRP is combinatorially large. It grows super-exponentially with the number of customers (due to partition–permutation coupling at the upper level), and for a fixed route, exponentially with the number of insertion points. Moreover, for each insertion point, the number of possible charging station options scales quadratically with the number of charging stations. This makes exact methods intractable even for moderately sized instances; hence, heuristic methods are not only preferable but arguably the only practical option for real-world problems.

\subsection{Objective Space Analysis}\label{subsec:objective-space-analysis}

In most bilevel programming problems, the leader's objective value can only be evaluated after solving the follower’s problem. However, in the bilevel model of the E-CVRP, each upper-level decision $x$ naturally defines a preliminary objective $\phi(x)$, representing the routing cost ignoring battery-related constraints, which is the objective of the classical CVRP subproblem. This structure allows partial evaluation of solution quality without solving the lower-level problem, providing valuable guidance for designing efficient search strategies. 

When a routing plan $x$ is given, the follower responds with the optimal charging strategy $y^*(x)$, resulting in the complete solution $(x, y^*(x))$. Intuitively, $\phi(x)$ should be generally positively correlated with the complete objective $F(x, y^*(x))$, since $y^*(x)$ aims to minimize additional detour distances imposed based on the given routing decision $x$. 

This correlation allows $\phi(x)$ to serve as a cheap surrogate for $F(x, y^*(x))$, especially valuable at early stages of the search process, as evaluating complete solutions is computationally expensive and not necessary. Therefore, when designing the algorithm, more search effort can be allocated to regions with lower $\phi(x)$ values to accelerate the search. For example, during early-stage convergence, only the routing solution $x$ is evaluated via $\phi(x)$; full evaluations of $(x, y^*(x))$ are deferred until $\phi(x)$ drops below a predefined threshold. 

However, it is worth noting that the correlation between $\phi(x)$ and $F(x, y^*(x))$ is not strictly monotonic, indicating a potential \emph{objective misalignment} phenomenon that needs both levels to be jointly optimized in a sophisticated way. 


\section{Bilevel Late Acceptance Hill Climbing}\label{sec:algorithm}

\subsection{Algorithm Overview}
\label{subsec:algorithm-overview}

The proposed b-LAHC algorithm operates through three key phases. In the \emph{initialization phase}, a greedy descent procedure ($\mathcal{G}$) first drives the upper-level routing solution $x$ to a local optimum. The \emph{exploration phase} then performs a systematic neighborhood search ($\mathcal{N}$) guided by a late acceptance mechanism, where the efficient follower optimizer ($\mathcal{F}_{\text{SE}}$) is conditionally activated to update charging configurations. Throughout this phase, the algorithm maintains a fixed-length history list ($\mathcal{P}$) and monitors search progress through the move acceptance ratio ($\rho_{\text{ns}}$). Finally, in the \emph{refinement phase}, upon termination, the exhaustive enumeration follower optimizer ($\mathcal{F}$) is invoked to polish the best-found solution. This three-phase approach effectively balances computational efficiency during the search with solution quality guarantees in the final output, while inheriting the theoretical convergence properties of the original LAHC framework that have been adapted to the bilevel problem structure. 

\begin{algorithm}[h]
\caption{Bilevel Late Acceptance Hill Climbing}
\label{alg:B-LAHC}
\renewcommand{\arraystretch}{0.8}
\begin{algorithmic}[1]
\Require \parbox[t]{0.9\linewidth}{
    Greedy descent $\mathcal{G}$, Neighborhood exploration $\mathcal{N}$, Follower optimizers $\mathcal{F}_{\text{SE}}$ and $\mathcal{F}$, 
    History list length $L_h$, Noise bounds $\alpha_{lb}$, $\alpha_{ub}$, 
    Max attempts $\eta_{\max}$, Follower activation threshold $\gamma$
}
\Ensure Best bilevel solution $(x^*, y^*)$ found

\State $F^* \gets +\infty$, \quad $x^* \gets \varnothing$, \quad $y^* \gets \varnothing$
\State Produce an initial upper-level solution $x$
\State Run greedy descent until local optimum $x \gets \mathcal{G}(x)$
\State $y_{\text{SE}}^* \gets \mathcal{F}_{\text{SE}}(x)$
\If{$F(x, y_{\text{SE}}^*) < F^*$}
    \State $(x^*, y^*) \gets (x, y_{\text{SE}}^*)$, $F^* \gets F(x, y_{\text{SE}}^*)$
\EndIf
\State Set $\phi^* \gets \phi(x)$, $I \gets 0$, $I_{\text{idle}} \gets 0$, $n \gets L_h$, $\rho_{n} \gets 1.0$
\For{$vi \gets 0$ \text{to} $L_h - 1$}
    \State $\mathcal{P}[vi] \gets \phi^* \cdot \textsc{UniformRandom}(\alpha_{lb}, \alpha_{ub})$
\EndFor  \Comment{$\phi_{vi}$ corresponds to $\mathcal{P}[vi]$}

\Repeat
    \State $x',\,\texttt{isMoved} \gets \mathcal{N}(x, \phi_{vi}, \eta_{\max})$
    \If{$\phi(x') < \phi(x)$}
        \State $I_{\text{idle}} \gets 0$, \quad $\phi^* \gets \min(\phi(x'), \phi^*)$
    \Else
        \State $I_{\text{idle}} \gets I_{\text{idle}} + 1$
    \EndIf

    \State $vi \gets I \bmod L_h$ \Comment{Virtual index in history list}
    \If{$vi = 0$}
        \State $\rho_{n} \gets n / L_h$, \quad $n \gets 0$
    \EndIf

    \If{$\texttt{isMoved}$}
        \State $x \gets x'$, $n \gets n + 1$, $\phi_{vi} \gets \min(\phi(x), \phi_{vi})$
        \If{$\phi(x) < \gamma \cdot \phi^*$}
            \State $y_{\text{SE}}^* \gets \mathcal{F}_{\text{SE}}(x)$
            \If{$F(x, y_{\text{SE}}^*) < F^*$}
                \State $(x^*, y^*) \gets (x, y_{\text{SE}}^*)$, $F^* \gets F(x, y_{\text{SE}}^*)$
            \EndIf
        \EndIf
    \EndIf

    \State $I \gets I + 1$
    \State $\texttt{hasConverged} \gets (I \ge 10^5 \,\textbf{and}\, I_{\text{idle}} \ge 0.02 \cdot I)$
    \Statex \phantom{$\texttt{hasConverged} \gets$} $\textbf{or} \, \rho_{n} \le 0.001$
    \State $\texttt{budgetExceeded} \gets \textsc{BudgetCheck}()$
\Until{\texttt{hasConverged} \textbf{or} \texttt{budgetExceeded}} 
\If{$\texttt{hasConverged}$ \textbf{and} $\neg \texttt{budgetExceeded}$}
    \State Restart the algorithm from Line~2 
\EndIf
\State $y^* \gets \mathcal{F}(x^*)$, \quad  $F^* \gets F(x^*, y^*)$ \Comment{Final refinement}
\State \Return $(x^*, y^*)$
\end{algorithmic}
\end{algorithm}

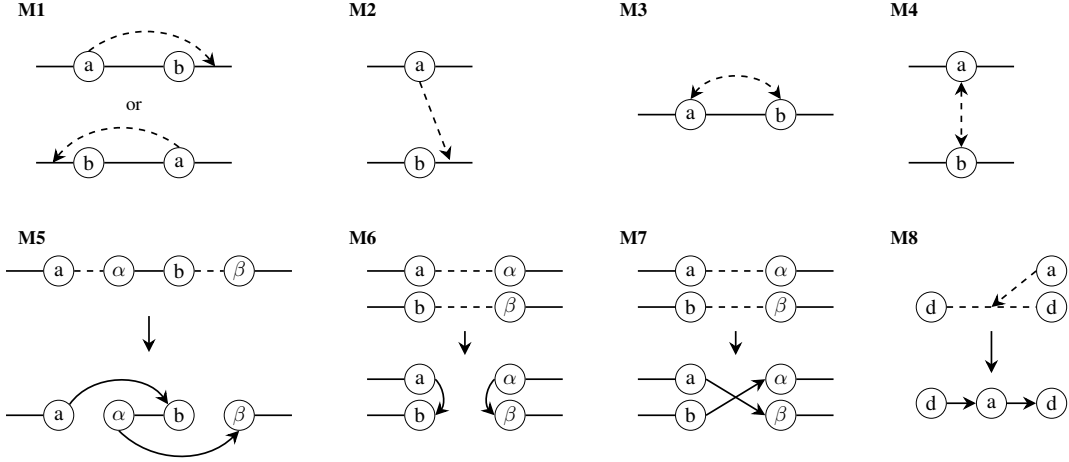
\begin{figure*}[!t]
\tikzset{
  vtx/.style={circle,draw,fill=white,minimum size=5.0mm,inner sep=0pt},
  lbl/.style={font=\small},
  seg/.style={line width=0.8pt},
  dseg/.style={line width=0.8pt,dashed},
  darr/.style={-{Stealth[length=2.2mm,width=2.2mm]},line width=0.8pt,dashed},
  ddarr/.style={-{Stealth[length=2.2mm,width=2.2mm]},line width=0.8pt},
  dbid/.style={<->,line width=0.8pt,dashed,>={Stealth[length=2.2mm,width=2.2mm]}},
  titlelbl/.style={font=\bfseries\small}
}

\centering
\scalebox{0.8}{%
\begin{tikzpicture}[x=1cm,y=1cm]

\def\dx{4.5}    
\def\dy{3.8}    
\def\yrowA{0}   
\def\yrowB{-\dy}

\newcommand{\vnode}[3]{\node[vtx] (#1) at #2 {#3};}

\begin{scope}[shift={(-1,\yrowA)}]
  \node[titlelbl,anchor=west] at (-1.3,1.75) {M1};

  \node[lbl] (tL) at (-1,0.8) {};
  \vnode{ta}{(0,0.8)}{a}
  \vnode{tb}{(1.5,0.8)}{b}
  \node[lbl] (tR) at (2.5,0.8) {};
  \draw[seg] (tL) -- (ta) -- (tb) -- (tR);
  \draw[darr,bend right=-45] (ta.north) to (2.1,0.8);

  \node[lbl] (bL) at (-1,-0.8) {};
  \vnode{bb}{(0,-0.8)}{b}
  \vnode{ba}{(1.5,-0.8)}{a}
  \node[lbl] (bR) at (2.5,-0.8) {};
  \draw[seg] (bL) -- (bb) -- (ba) -- (bR);
  \draw[darr,bend left=-45] (ba.north) to (-0.6,-0.8);

  \node[lbl] at (0.75,0.15) {or};
\end{scope}

\begin{scope}[shift={(\dx,\yrowA)}]
  \node[titlelbl,anchor=west] at (-1.3,1.75) {M2};

  \node[lbl] (L1) at (-1,0.8) {};
  \vnode{a2}{(0,0.8)}{a}
  \node[lbl] (R1) at (1.0,0.8) {};
  \draw[seg] (L1) -- (a2) -- (R1);

  \node[lbl] (L2) at (-1,-0.8) {};
  \vnode{b2}{(0,-0.8)}{b}
  \node[lbl] (R2) at (1.0,-0.8) {};
  \draw[seg] (L2) -- (b2) -- (R2);

  \draw[darr] (a2.south) -- ++(0.5,-1.3);
\end{scope}

\begin{scope}[shift={(2*\dx,\yrowA)}]
  \node[titlelbl,anchor=west] at (-1.3,1.75) {M3};

  \node[lbl] (L3) at (-1,0) {};
  \vnode{a3}{(0,0)}{a}
  \vnode{b3}{(1.5,0)}{b}
  \node[lbl] (R3) at (2.5,0) {};
  \draw[seg] (L3) -- (a3) -- (b3) -- (R3);
  \draw[dbid,bend right=-55] (a3.north) to (b3.north);
\end{scope}

\begin{scope}[shift={(3*\dx,\yrowA)}]
  \node[titlelbl,anchor=west] at (-1.3,1.75) {M4};

  \node[lbl] (L41) at (-1,0.8) {};
  \vnode{a4}{(0,0.8)}{a}
  \node[lbl] (R41) at (1.0,0.8) {};
  \draw[seg] (L41) -- (a4) -- (R41);

  \node[lbl] (L42) at (-1,-0.8) {};
  \vnode{b4}{(0,-0.8)}{b}
  \node[lbl] (R42) at (1.0,-0.8) {};
  \draw[seg] (L42) -- (b4) -- (R42);

  \draw[dbid] (a4.south) -- (b4.north);
\end{scope}

\begin{scope}[shift={(-1,\yrowB)}]
  \node[titlelbl,anchor=west] at (-1.3,1.75) {M5};

  \node[lbl] (tL5) at (-1.5,1.2) {};
  \vnode{ta5}{(-0.5,1.2)}{a}
  \vnode{tp5}{(0.5,1.2)}{$\alpha$}
  \vnode{tb5}{(1.5,1.2)}{b}
  \vnode{tq5}{(2.5,1.2)}{$\beta$}
  \node[lbl] (tR5) at (3.5,1.2) {};
  \draw[seg] (tL5) -- (ta5);
  \draw[seg] (tp5) -- (tb5);
  \draw[seg] (tq5) -- (tR5);
  \draw[dseg] (ta5) -- (tp5);
  \draw[dseg] (tb5) -- (tq5);

  \node[lbl] (bL5) at (-1.5,-1.2) {};
  \vnode{ba5}{(-0.5,-1.2)}{a}
  \vnode{bb5}{(1.5,-1.2)}{b}
  \vnode{bp5}{(0.5,-1.2)}{$\alpha$}
  \vnode{bq5}{(2.5,-1.2)}{$\beta$}
  \node[lbl] (bR5) at (3.5,-1.2) {};
  \draw[seg] (bL5) -- (ba5);
  \draw[seg] (bp5) -- (bb5);
  \draw[seg] (bq5) -- (bR5);

  \draw[ddarr,bend right=-55] (ba5.north east) to (bb5.north west);
  \draw[ddarr,bend right=45] (bp5.south) to (bq5.south);

  \draw[ddarr] (1,0.45) -- (1,-0.15);
\end{scope}

\begin{scope}[shift={(1*\dx,\yrowB)}]
  \node[titlelbl,anchor=west] at (-1.3,1.75) {M6};

  \node[lbl] (t1L6) at (-1,1.2) {};
  \vnode{ta6}{(0,1.2)}{a}
  \vnode{tp6}{(1.5,1.2)}{$\alpha$}
  \node[lbl] (t1R6) at (2.5,1.2) {};
  \draw[seg] (t1L6) -- (ta6);
  \draw[seg] (tp6) -- (t1R6);
  \draw[dseg] (ta6) -- (tp6);

  \node[lbl] (t3L6) at (-1,0.6) {};
  \vnode{tb6}{(0,0.6)}{b}
  \vnode{tq6}{(1.5,0.6)}{$\beta$}
  \node[lbl] (t3R6) at (2.5,0.6) {};
  \draw[seg] (t3L6) -- (tb6);
  \draw[seg] (tq6) -- (t3R6);
  \draw[dseg] (tb6) -- (tq6);

  \node[lbl] (b1L6) at (-1,-0.6) {};
  \vnode{ba6}{(0,-0.6)}{a}
  \vnode{bp6}{(1.5,-0.6)}{$\alpha$}
  \node[lbl] (b1R6) at (2.5,-0.6) {};
  \draw[seg] (b1L6) -- (ba6);
  \draw[seg] (bp6) -- (b1R6);

  \node[lbl] (b3L6) at (-1,-1.2) {};
  \vnode{bb6}{(0,-1.2)}{b}
  \vnode{bq6}{(1.5,-1.2)}{$\beta$}
  \node[lbl] (b3R6) at (2.5,-1.2) {};
  \draw[seg] (b3L6) -- (bb6);
  \draw[seg] (bq6) -- (b3R6);

  \draw[ddarr,bend left=45] (ba6.east) to (bb6.east);
  \draw[ddarr,bend left=-45]  (bp6.west) to (bq6.west);

  \draw[ddarr] (0.75,0.2) -- (0.75,-0.2);
\end{scope}

\begin{scope}[shift={(2*\dx,\yrowB)}]
  \node[titlelbl,anchor=west] at (-1.3,1.75) {M7};

  \node[lbl] (t1L7) at (-1,1.2) {};
  \vnode{ta7}{(0,1.2)}{a}
  \vnode{tp7}{(1.5,1.2)}{$\alpha$}
  \node[lbl] (t1R7) at (2.5,1.2) {};
  \draw[seg] (t1L7) -- (ta7);
  \draw[seg] (tp7) -- (t1R7);
  \draw[dseg] (ta7) -- (tp7);

  \node[lbl] (t3L7) at (-1,0.6) {};
  \vnode{tb7}{(0,0.6)}{b}
  \vnode{tq7}{(1.5,0.6)}{$\beta$}
  \node[lbl] (t3R7) at (2.5,0.6) {};
  \draw[seg] (t3L7) -- (tb7);
  \draw[seg] (tq7) -- (t3R7);
  \draw[dseg] (tb7) -- (tq7);

  \node[lbl] (b1L7) at (-1,-0.6) {};
  \vnode{ba7}{(0,-0.6)}{a}
  \vnode{bp7}{(1.5,-0.6)}{$\alpha$}
  \node[lbl] (b1R7) at (2.5,-0.6) {};
  \draw[seg] (b1L7) -- (ba7);
  \draw[seg] (bp7) -- (b1R7);

  \node[lbl] (b3L7) at (-1,-1.2) {};
  \vnode{bb7}{(0,-1.2)}{b}
  \vnode{bq7}{(1.5,-1.2)}{$\beta$}
  \node[lbl] (b3R7) at (2.5,-1.2) {};
  \draw[seg] (b3L7) -- (bb7);
  \draw[seg] (bq7) -- (b3R7);

  \draw[ddarr] (ba7.east) to (bq7.west);
  \draw[ddarr]  (bb7.east) to (bp7.west);

  \draw[ddarr] (0.75,0.2) -- (0.75,-0.2);
\end{scope}

\begin{scope}[shift={(3*\dx,\yrowB)}]
  \node[titlelbl,anchor=west] at (-1.3,1.75) {M8};

  \vnode{ta8}{(1.5,1.2)}{a}
  
  \vnode{tb8}{(-0.5,0.6)}{d}
  \vnode{tq8}{(1.5,0.6)}{d}
  \draw[dseg] (tb8) -- (tq8);
  \draw[darr,bend left=-45] (ta8.west) -- (0.5, 0.6);

  \vnode{bdo8}{(-0.5,-1)}{d}
  \vnode{ba8}{(0.5,-1)}{a}
  \vnode{bdi8}{(1.5,-1)}{d}
  \draw[ddarr] (bdo8) -- (ba8);
  \draw[ddarr] (ba8) -- (bdi8);

  \draw[ddarr] (0.5,0.2) -- (0.5,-0.5);
\end{scope}
\end{tikzpicture}
}
\caption{The eight move operators illustrated by simplified route segments.}
    \label{fig:moves}
\end{figure*}

Algorithm~\ref{alg:B-LAHC} details b-LAHC procedure. An initial solution $x$ is generated via random customer permutation and the split method~\cite{prins2014order} (Line~2), followed by greedy descent $\mathcal{G}$ to reach a local optimum (Line~3). The efficient follower optimizer $\mathcal{F}_{\text{SE}}$ is applied to compute the optimal lower-level decision $y_{\text{SE}}^*(x)$ using the simple enumeration (SE) method (Line~4; see Section~\ref{subsec:lower-level-decision}), and the best-known solution $(x^*, y^*)$ and the corresponding objective $F^*$ are updated if improved (Lines~5--7). Next, the algorithm initializes the best-known surrogate cost $\phi^*$, the iteration counter $I$, the idle iteration counter $I_{\text{idle}}$, the number of successful moves $n$ (i.e., the moves within the current history list that accept the candidate solution), and the corresponding success ratio $\rho_{n}$ (Line~8). From among these, $\rho_{n}$ serves as an additional indicator of search progress, added to the original LAHC scheme. It is updated periodically and enhances convergence detection. Subsequently, a history list $\mathcal{P}$ of length $L_h$ is created to store past surrogate costs for the late acceptance criterion. Each entry is initialized by applying noise to $\phi^*$, multiplying it by a random factor between $\alpha_{\text{lb}}$ (slightly below 1.0) and $\alpha_{\text{ub}}$ (slightly above 1.0). This balanced perturbation avoids overly strict or lenient acceptance thresholds during the search (Lines~9--11). 

At each iteration, $\mathcal{N}(x, \phi_{vi}, \eta_{\max})$ is invoked, yielding a candidate solution $x'$ and a flag \texttt{isMoved} that records whether the move is accepted (Line~13). If $\phi(x')$ improves upon $\phi(x)$, the idle counter $I_{\text{idle}}$ is reset to 0 and $\phi^*$ is updated; otherwise, $I_{\text{idle}}$ is incremented by 1 (Lines~14--18). The virtual index $vi$ in the history list is then updated (Line~19). Whenever the history list $\mathcal{P}$ completes a cycle, $\rho_{n}$ is recomputed and $n$ is reset (Lines~20--22). If the flag \texttt{isMoved} is true, the update process is triggered (Lines~23--31): $x$ is replaced by $x'$, $n$ is incremented, and the historical value $\phi_{vi}$ is updated (Line~24). To avoid unnecessary lower-level evaluations and to concentrate the search efforts on promising regions, as discussed in Section~\ref{subsec:objective-space-analysis}, the follower is invoked only when $\phi(x)$ is close enough to the best surrogate cost $\phi^*$ found so far (Line~25). Once the follower is triggered, the optimizer $\mathcal{F}_{\text{SE}}$ computes the optimal lower-level decision $y_{\text{SE}}^*(x)$, and the best-known solution $(x^*, y^*)$ together with its objective $F^*$ are updated whenever an improvement is found (Lines~26--29). At the end of each iteration, $I$ is incremented (Line~32), and two termination flags are updated: 
\begin{itemize}[leftmargin=1em, itemsep=0pt, topsep=0pt, parsep=0pt]
  \item \texttt{hasConverged}: set to true when either (i) excessive idle iterations are detected ($I \ge 10^5$ and $I_{\text{idle}} \ge 0.02I$) or (ii) the search success ratio deteriorates ($\rho_{n} \le 0.001$).
  \item \texttt{budgetExceeded}: set to true when the computational budget is exhausted.
\end{itemize}
The current loop terminates when either condition is met (Line~35). If convergence is detected before the budget is exhausted, the algorithm restarts from Line~2 and continues until the next termination check (Lines~36--38). 

Finally, before returning the solution, a final refinement step is performed: the exhaustive follower $\mathcal{F}$ is invoked to compute the optimal charging configuration $y^*(x^*)$ for the best-found routing plan $x^*$ (Line~39), ensuring that the returned bilevel solution $(x^*, y^*)$ represents a verified optimum (Line~40).

\subsection{Move Operators}
\label{subsec:move-operators}

Move operators define how a current solution $x$ is transformed into a neighboring solution $x'$ by modifying part of $x$. Both greedy descent ($\mathcal{G}$) and neighborhood exploration ($\mathcal{N}$) are built upon these operators.

Inspired by Prins~\cite{prins2004simple}, eight move operators are adopted to modify the upper-level routing solution $x$. Let $a$ and $b$ be two distinct customer nodes, which may belong to the same route or different routes. Let $\alpha$ and $\beta$ denote the immediate successors of $a$ and $b$, respectively. The eight operators are illustrated in Fig.~\ref{fig:moves} and summarized below:
\begin{itemize}
  \item[M1.] Relocate $a$ within the same route by inserting it before or after $b$.
  \item[M2.] Relocate $a$ from its current route and insert it after $b$ in another route.
  \item[M3.] Swap $a$ and $b$ in the same route.
  \item[M4.] Swap $a$ and $b$ between two different routes.
  \item[M5.] In the same route, replace arcs $(a,\alpha)$ and $(b,\beta)$ with $(a,b)$ and $(\alpha,\beta)$.
  \item[M6.] In two different routes, replace arcs $(a,\alpha)$ and $(b,\beta)$ with $(a,b)$ and $(\alpha,\beta)$.
  \item[M7.] In two different routes, replace arcs $(a,\alpha)$ and $(b,\beta)$ with $(a,\beta)$ and $(b,\alpha)$.
  \item[M8.] Remove $a$ from its current route and insert it into an empty route.
\end{itemize}

To apply an operator, a target $T$ is first selected from the current solution $x$, where $T$ can be either a single route or a pair of routes. From $T$, two distinct customer nodes $(a,b)$ are chosen. The selected operator $op$ is subsequently applied, yielding a neighboring solution $x'$. This process is formally expressed as $x' = op(x, T, a, b)$.

These operators can be categorized into three groups: (i) intra-route moves (M1, M3, M5), (ii) inter-route moves (M2, M4, M6, M7), and (iii) inter-route moves, involving empty routes (M8). M8 differs from the others in that it moves a node to an empty route. Only M8 can increase the total number of routes in the solution $x$, whereas M1–M7 either reduce or preserve the current route count. This operator is designed to handle special cases where using more vehicles may result in a lower total cost. M8 facilitates the exploration of more diverse regions in the solution space and helps the search to escape from local optima.

\subsection{Greedy Descent}
\label{subsec:greedy-descent}

The greedy descent procedure $\mathcal{G}(x)$ performs iterative improvement on $x$ until reaching a local optimum $x^{\text{opt}}$, i.e., $\mathcal{G}(x)$: $x \mapsto x^{\text{opt}}$. This computationally efficient exploitation rapidly reduces $\phi(x)$, directing the search toward promising regions of the search space.

\begin{algorithm}[!h]
	\caption{Greedy Descent $\mathcal{G}$}
	\label{alg:greedy-descent}
    \renewcommand{\arraystretch}{0.8}
	\begin{algorithmic}[1]
		\Require Current solution $x$, set of move operators $\mathcal{I}$
		\Ensure A locally optimal solution $x^{\text{opt}}$
		\Repeat
		\State $\texttt{aImp} \gets \text{False}$, \textsc{Shuffle}$(\mathcal{I})$
		\For{each operator $op \in \mathcal{I}$}
		\State $\texttt{oImp} \gets \text{False}$
		\For{each target $T \in x$} \Comment{Enumerate}
		\Repeat
		\State $\texttt{mDone} \gets \text{False}$
		\For{each $(a,b) \in T$} \Comment{Enumerate}
		\State $x' \gets op(x, T, a, b)$
		\If{$\phi(x') < \phi(x)$} 
		\State $x \gets x'$
		\State $\texttt{mDone} \gets \text{True}$, $\texttt{oImp} \gets \text{True}$
		\State \textbf{break} 
		\EndIf
		\EndFor
		\State \textsc{UpdateTargetInfo}$(x)$ 
		\Until{$\texttt{mDone} = \text{False}$}
		\EndFor
		\State $\texttt{aImp} \gets \texttt{aImp} \lor \texttt{oImp}$
		\EndFor 
		\Until{$\texttt{aImp} = \text{False}$} 
		\State \Return $x^{\text{opt}} \gets x$
	\end{algorithmic}
\end{algorithm}

Algorithm~\ref{alg:greedy-descent} performs a multi-level best-improvement descent over an operator set $\mathcal{I}$ (excluding M8). Each outer iteration resets the any-improvement flag $\texttt{aImp}$ and shuffles $\mathcal{I}$ to avoid ordering bias (Line~2). For each operator $op \in \mathcal{I}$, the operator-level flag $\texttt{oImp}$ is cleared (Line~4). Then, for every target $T$ (a route or route pair) in the current solution $x$, the algorithm repeatedly scans pairs $(a,b) \in T$ and applies $op$ to generate $x'$ (Lines~5--18). If $\phi(x') < \phi(x)$, it updates $x$ and sets both $\texttt{mDone}$ (move-applied flag) and $\texttt{oImp}$; otherwise it continues (Lines~8--15). At the end of each attempt, information about the affected route or route pair is updated to reflect the latest structural changes in $x$ (Line~16). This step is specifically for inter-route moves as they may create empty routes and thus change the subsequent targets in $x$ (affecting Line~5). The per-target loop stops when no further improving move is found.  

After all targets for the current operator are processed, $\texttt{aImp}$ is updated (Line~19). The outer loop repeats while any improvement exists. Once no operator yields further improvement, the procedure terminates and returns the locally optimal solution $x^{\text{opt}}$.

\subsection{Neighborhood Exploration}
\label{subsec:neighbourhood-exploration}

The neighborhood exploration component drives the upper-level solution $x$ through iterative transitions across its solution space, guided by the late acceptance mechanism. 

Algorithm~\ref{alg:neighbourhood-exploration} details this procedure. The process begins by initializing the candidate solution $x'$ and the move flag (Line~1). The attempt counter $i$ is initialized (Line~2), and a move operator $op$ is selected uniformly at random (u.a.r.) from the operator set $\mathcal{M}$ (Line~3). $\mathcal{M}$ includes all operators described in Section~\ref{subsec:move-operators}. 

\begin{algorithm}[!h]
	\caption{Neighborhood Exploration $\mathcal{N}$}
	\label{alg:neighbourhood-exploration}
    \renewcommand{\arraystretch}{0.8}
	\begin{algorithmic}[1]
		\Require \parbox[t]{0.92\linewidth}{
			Current solution $x$, history solution cost $\phi_{vi}$, \\
			maximum attempts $\eta_{\max}$, move operators $\mathcal{M}$
		}
		\Ensure A neighbour solution $x'$, a flag $\texttt{isMoved}$
		
		\State $x' \gets x$, $\texttt{isMoved} \gets \text{false}$
		\State $i \gets 0$ \Comment{Attempt counter}
		\State Randomly select a move operator $op \in \mathcal{M}$
		
		\Repeat
		\State Randomly select a target $T \in x$ 
		\State Randomly select a node $a \in T$
		\State Enumerate candidate positions $b$ based on $op$
		\For{each feasible $b$}
		\State Generate $x'' \gets op(x, T, a, b)$
		\If{$\phi(x'') < \phi_{vi} \ \textbf{or} \ \phi(x'') < \phi(x)$}
		\State $x' \gets x''$, $\texttt{isMoved} \gets \text{true}$
		\State \textbf{break}
		\EndIf
		\EndFor
		\State $i \gets i + 1$
		\Until{$\texttt{isMoved} \ \textbf{or} \ i \ge \eta_{\max}$}
		\State \Return $x'$, $\texttt{isMoved}$
	\end{algorithmic}
\end{algorithm}

At each attempt, a target $T$ is u.a.r. chosen from the current solution $x$ (Line~5). From $T$, a customer node $a$ is selected (Line~6), and the candidate positions $b$ are enumerated based on the chosen operator $op$ (Line~7). For operator M8, $b$ refers to the depot node of an empty route.

For each feasible position $b$ (Line~8), a candidate neighbour solution $x''$ is generated by applying the operator (Line~9). The late acceptance condition then checks if $x''$ is better than either $\phi(x)$ or the historical value $\phi_{vi}$ (Line~10). Once satisfied, $x''$ is accepted as the new neighbour $x'$, $\texttt{isMoved}$ is set to \text{true} (Line~11). The process terminates early once the condition is satisfied or after reaching the maximum number of attempts $\eta_{\max}$ (Line~16).

\subsection{Lower-level Decision}
\label{subsec:lower-level-decision}

The lower-level charging decision involves optimally inserting charging stations along a fixed route. This problem is referred to as the Fixed Route Vehicle Charging Problem (FRVCP), which has been proven to be \NPhard~\cite{montoya2017electric}. The lower-level decision space $\mathcal{Y}(x)$ is fully determined by the upper-level routing solution $x$, as introduced in Section~\ref{subsec:solution-space-analysis}. Within $\mathcal{Y}(x)$, all potential charging insertion points, i.e., the gaps between consecutive nodes, along each route are considered simultaneously. Moreover, multiple possible charging configurations may exist for each insertion point. 

To reduce the complexity of the lower-level decision, we bound the number of charging station visits on each route $R_v$ between $\ell b_{v}$ and $\ell b_{v}+1$, where $\ell b_{v}$ denotes the minimum required visits for $R_v$. It is calculated by dividing the total travel distance of the route $\phi(R_v)$ by the maximum cruising range of an EV at full charge $Q_b/h$:
\begin{equation}
\ell b_{v} = \left\lfloor \frac{\phi(R_v)}{Q_b/h} \right\rfloor \notag
\label{eq:lower-bound-charging}
\end{equation}

Under this restriction, we exhaustively enumerate all feasible configurations with up to two consecutive charging station visits at each insertion point. The resulting optimal charging configuration is denoted as $y^*$, and the exhaustive enumeration procedure is referred to as $\mathcal{F}$. However, since $\mathcal{F}$ is computationally expensive, it would be impractical to include it throughout the search. 

Thus, an efficient lower-level optimization alternative is essential. Specifically, we adopt the Simple Enumeration (SE) method proposed in~\cite{jia2022confidence}, which significantly reduces computational complexity by restricting the search to configurations with at most one charging station per insertion point and precomputing the ``best charging station'' for every pair of customers (or depot-customer pairs). This best charging station is selected to minimize the additional travel distance incurred by the detour, as defined below:
\begin{equation}
\theta_{ij} = \arg\min_{\theta \in \mathcal{V}_f} \left( d_{i\theta} + d_{\theta j} \right), \quad \forall i, j \in \mathcal{V}_c \cup \{d\},\ i \ne j  \notag
\label{eq:best-station}
\end{equation}

Under SE, the number of configurations per route is determined by selecting either $\ell b_{v}$ or $\ell b_{v} + 1$ insertion points from the $L_v + 1$ available gaps, with the optimal charging station at each chosen point fixed in advance: 
$$
\binom{L_v + 1}{\ell b_{v}} + \binom{L_v + 1}{\ell b_{v} + 1}
$$
where $L_v + 1$ is the number of insertion points. The total complexity is:
$$
|\mathcal{Y}_{\text{SE}}(x)| = \prod_{v=1}^{M} \left[ \binom{L_v + 1}{\ell b_v} + \binom{L_v + 1}{\ell b_v + 1} \right]
$$
which is polynomial in practice, since $\ell b_v$ is typically small (in most cases, an EV requires no more than three charging stops per trip), and $L_v$ grows linearly with the size of the problem. SE thus reduces the theoretically exponential complexity of $\mathcal{Y}(x)$ to a tractable scale even for large-scale instances. We denote the optimal charging configuration returned by SE as $y_{\text{SE}}^*$, and the SE process as $\mathcal{F}_{\text{SE}}$.

In Algorithm~\ref{alg:B-LAHC}, $\mathcal{F}_{\text{SE}}$ serves as the default efficient optimizer, whereas $\mathcal{F}$ is reserved for final solution refinement to ensure optimality, thereby achieving a balance between computational efficiency and solution quality.

\begin{table}[!t]
\centering
\caption{Details of the IEEE WCCI-2020 benchmark set}
\label{tab:instance}
\scriptsize
\setlength{\tabcolsep}{3pt}
\setlength\tabcolsep{4.0pt}  
\renewcommand{\arraystretch}{0.8}
\begin{tabular}{lccccccc}
\toprule
name & $|\mathcal{V}_c|$ & $|\mathcal{V}_f|$ & $M$ & $Q_c$ & $Q_b$ & $h$ & $UB$ \\
\midrule
E22  & 21  & 8  & 4   & 6000 & 94   & 1.2 & 384.67 \\
E23  & 22  & 9  & 3   & 4500 & 190  & 1.2 & 573.13 \\
E30  & 29  & 6  & 4   & 4500 & 178  & 1.2 & 511.25 \\
E33  & 32  & 6  & 4   & 8000 & 209  & 1.2 & 869.89 \\
E51  & 50  & 5  & 5   & 160  & 105  & 1.2 & 570.17 \\
E76  & 75  & 7  & 7   & 220  & 98   & 1.2 & 723.36 \\
E101 & 100 & 9  & 8   & 200  & 103  & 1.2 & 899.88 \\
\midrule  
X143 & 142 & 4  & 7   & 1190 & 2243 & 1.0 & -- \\
X214 & 213 & 9  & 11  & 944  & 987  & 1.0 & -- \\
X351 & 350 & 35 & 40  & 436  & 649  & 1.0 & -- \\
X459 & 458 & 20 & 26  & 1106 & 929  & 1.0 & -- \\
X573 & 572 & 6  & 30  & 210  & 1691 & 1.0 & -- \\
X685 & 684 & 25 & 75  & 408  & 911  & 1.0 & -- \\
X749 & 748 & 30 & 98  & 396  & 790  & 1.0 & -- \\
X819 & 818 & 25 & 171 & 358  & 926  & 1.0 & -- \\
X916 & 915 & 32 & 207 & 33   & 1591 & 1.0 & -- \\
X1001 & 1000 & 9 & 43 & 131  & 1684 & 1.0 & -- \\
\bottomrule
\end{tabular}
\end{table}
 
\section{Experiments}
\label{sec:experiments}

\begin{table}[!t]
\centering
\caption{Overview of Methods for the WCCI-2020 EVRP Benchmark}
\label{tab:method-summary}
\scriptsize
\setlength{\tabcolsep}{3pt}
\renewcommand{\arraystretch}{0.8}
\begin{tabular}{llll}
\toprule
\textbf{Methods} & \textbf{Stop Criteria} & \textbf{Language/Tool} & \textbf{Code Available} \\
\midrule
MILP~\cite{mavrovouniotis2020techreport}     & Max Evals & Gurobi   & No \\
VNS~\cite{woller2025variable}              & Max Evals & C++      &Yes \\
SA                  & Max Evals & C++      & No \\
GA~\cite{hien2023greedy}       & Max Evals & C++      & Yes \\
HHASA-TS~\cite{rodriguez2024new} & Max Evals & MATLAB   & Yes \\
BACO~\cite{jia2021bilevel}     & Max Time  & C++      & Yes \\
CBACO-I~\cite{jia2022confidence}  & Max Time  & C++      & Yes \\
TAMLS~\cite{chen2024efficient}    & Max Time  & C++      & No \\
CBMA~\cite{qin2024confidence}     & Max Time  & C++      & Yes \\
b-LAHC                & Both      & C++      & Yes \\
\bottomrule
\end{tabular}
\end{table}

\begin{table}[!t]
\centering
\caption{Main parameters of the b-LAHC algorithm}
\label{tab:parameters}
\renewcommand{\arraystretch}{0.8}
\begin{tabular}{llc}
\toprule
\textbf{Parameter} & \textbf{Description} & \textbf{Value} \\
\midrule
$L_h$ & History length                                & 5723 \\
$\eta_{\max}$  & Maximum attempts                      & 60   \\
$\gamma$ & Follower activation threshold              & 1.01 \\
\bottomrule
\end{tabular}
\end{table}

\begin{table}[!t]
\centering
\caption{Rank correlation and Top-$k$\% set overlap between $\phi(x)$ and $F(x,y^*(x))$.}
\label{tab:kendall-recall}
\scriptsize
\setlength{\tabcolsep}{1pt}
\renewcommand{\arraystretch}{0.8}
\begin{tabular}{lcccccc}
\toprule
\textbf{Instance} & \textbf{\#Samples} & $\boldsymbol{\tau_b}$ & \textbf{Recall@1\%} & \textbf{Recall@5\%} & \textbf{Recall@10\%} & \textbf{Recall@20\%} \\
\midrule
E22  & 60,163 & 0.9245 & 0.7425 & 0.8136 & 0.8764 & 0.9253 \\
E23  & 100,717 & 0.9646 & 0.8978 & 0.9255 & 0.9387 & 0.9646 \\
E30  & 182,830 & 0.9750 & 0.9043 & 0.9354 & 0.9567 & 0.9745 \\
E33  & 115,460 & 0.9529 & 0.8823 & 0.9531 & 0.9671 & 0.9792 \\
E51  & 391,533 & 0.9802 & 0.7439 & 0.9279 & 0.9631 & 0.9814 \\
E76  & 712,352 & 0.9755 & 0.7650 & 0.8990 & 0.9157 & 0.9676 \\
E101 & 854,090 & 0.9750 & 0.7685 & 0.9201 & 0.9478 & 0.9743 \\
X143 & 1,354,989 & 0.9812 & 0.9728 & 0.9341 & 0.9542 & 0.9792 \\
X214 & 3,752,720 & 0.9785 & 0.5141 & 0.8491 & 0.9423 & 0.9857 \\
X351 & 12,244,328 & 0.9877 & 0.8522 & 0.9558 & 0.9702 & 0.9876 \\
X459 & 8,990,056 & 0.9903 & 0.9291 & 0.9639 & 0.9902 & 0.9924 \\
X573 & 12,317,882 & 0.9811 & 0.0278 & 0.8562 & 0.9429 & 0.9846 \\
X685 & 31,464,639 & 0.9869 & 0.4730 & 0.9087 & 0.9630 & 0.9879 \\
X749 & 46,890,428 & 0.9867 & 0.5526 & 0.9034 & 0.9653 & 0.9848 \\
X819 & 61,717,308 & 0.9661 & 0.3169 & 0.6893 & 0.8790 & 0.9616 \\
X916 & 78,677,293 & 0.9666 & 0.4708 & 0.6854 & 0.8890 & 0.9620 \\
X1001 & 27,402,846 & 0.9889 & 0.6183 & 0.9256 & 0.9671 & 0.9904 \\
\bottomrule
\end{tabular}
\end{table}

\begin{table*}[!t]
\centering
\caption{Objective value comparison with state-of-the-art algorithms using the Max Evals stop criterion}
\label{tab:max-evals}
\scriptsize
\setlength{\tabcolsep}{5pt}
\renewcommand{\arraystretch}{0.8}
\begin{tabular}{llcccccccccc}
\toprule
Instance & Index & BKS & VNS  & SA & GA  & HHASA-TS & BACO & CBACO-I & CBMA & b-LAHC & Gap \\
\midrule
\multirow{3}{*}{E22} & best & 384.67 & \textbf{384.67} & \textbf{384.67} & \textbf{384.67} & \textbf{384.67} & \textbf{384.67} & \textbf{384.67} & \textbf{384.67} & \textbf{384.67} & -- \\
 & mean & 384.67 & \textbf{384.67} & \textbf{384.67} & \textbf{384.67} & \textbf{384.67} & \textbf{384.67} & \textbf{384.67} & \textbf{384.67} & 385.25 & 0.15\% \\
 & std. & -- & 2.11 & 0.00 & 0.00 & 0.00 & 0.00 & 0.00 & 0.00 & 0.28 & -- \\
\midrule
\multirow{3}{*}{E23} & best & 571.94 & \textbf{571.94} & \textbf{571.94} & \textbf{571.94} & \textbf{571.94} & \textbf{571.94} & \textbf{571.94} & \textbf{571.94} & \textbf{571.94} & -- \\
 & mean & 571.94 & \textbf{571.94} & \textbf{571.94} & \textbf{571.94} & \textbf{571.94} & \textbf{571.94} & \textbf{571.94} & \textbf{571.94} & \textbf{571.94} & 0.00\% \\
 & std. & -- & 0.00 & 0.00 & 0.00 & 0.00 & 0.00 & 0.00 & 0.00 & 0.00 & --  \\
\midrule
\multirow{3}{*}{E30} & best & 509.47 & \textbf{509.47} & \textbf{509.47} & \textbf{509.47} & \textbf{509.47} & \textbf{509.47} & \textbf{509.47} & \textbf{509.47} & \textbf{509.47} & -- \\
 & mean & 509.47 & \textbf{509.47} & \textbf{509.47} & \textbf{509.47} & \textbf{509.47} & \textbf{509.47} & \textbf{509.47} & \textbf{509.47} & \textbf{509.47} & 0.00\%  \\
 & std. & -- & 0.00 & 0.00 & 0.00 & 0.00 & 0.00 & 0.00 & 0.00 & 0.00 & -- \\
\midrule
\multirow{3}{*}{E33} & best & 840.14 & \textbf{840.14} & 840.57 & 844.25 & \textbf{840.14} & 846.02 & 846.96 & 840.57 & \textbf{840.14} & -- \\
 & mean & 840.43 & 840.43 & 854.07 & 845.62 & 840.70 & 847.20 & 847.74 & 840.57 & \textbf{840.14} & -0.03\% \\
 & std. & -- & 1.18 & 12.80 & 0.92 & 1.40 & 1.21 & 0.56 & 0.00 & 0.00 & -- \\
\midrule
\multirow{3}{*}{E51} & best & 529.90 & \textbf{529.90} & 533.66 & \textbf{529.90} & \textbf{529.90} & 546.85 & \textbf{529.90} & \textbf{529.90} & \textbf{529.90} & -- \\
 & mean & 529.90 & 543.26 & 533.66 & 542.08 & 536.98 & 555.48 & 535.39 & \textbf{529.90} & 531.19 & 0.24\% \\
 & std. & -- & 3.52 & 0.00 & 8.57 & 7.27 & 4.47 & 4.72 & 0.00 & 3.59 & -- \\
\midrule
\multirow{3}{*}{E76} & best & 692.64 & \textbf{692.64} & 701.03 & 697.27 & 692.74 & 743.20 & 726.02 & \textbf{692.64} & 696.27 & -- \\
 & mean & 694.61 & 697.89 & 712.17 & 717.30 & 694.96 & 757.08 & 735.34 & \textbf{694.61} & 699.54 & 0.71\%  \\
 & std. & -- & 3.09 & 5.78 & 9.58 & 1.63 & 5.79 & 4.85 & 2.10 & 2.91 & -- \\
\midrule
\multirow{3}{*}{E101} & best & 837.10 & 839.29 & 845.84 & 852.69 & \textbf{837.10} & 905.09 & 905.09 & 841.80 & 840.55 & -- \\
 & mean & 843.10 & 853.34 & 852.48 & 872.69 & \textbf{843.10} & 905.09 & 905.09 & 844.96 & 844.01 & 0.11\%  \\
 & std. & -- & 4.73 & 3.44 & 9.58 & 3.90 & 0.00 & 0.00 & 2.63 & 3.16 & -- \\
\midrule
\multirow{3}{*}{X143} & best & 15910.86 & 16028.05 & 16610.37 & 16488.60 & \textbf{15910.86} & 17407.82 & 17407.82 & 16288.88 & 15916.19 & -- \\
 & mean & 16214.37 & 16459.31 & 17188.90 & 16911.50 & 16214.37 & 17407.82 & 17407.82 & 16489.90 & \textbf{16103.45} & -0.68\%  \\
 & std. & -- & 242.59 & 170.44 & 282.30 & 215.77 & 0.00 & 0.00 & 142.22 & 85.93 & -- \\
\midrule
\multirow{3}{*}{X214} & best & 11090.28 & 11323.56 & 11404.44 & 11762.07 & \textbf{11090.28} & 12267.88 & 12267.88 & 11461.70 & 11114.46 & -- \\
 & mean & 11206.60 & 11482.20 & 11680.35 & 12007.06 & \textbf{11206.60} & 12267.88 & 12267.88 & 11608.98 & 11245.67 & 0.35\%  \\
 & std. & -- & 76.14 & 116.47 & 156.69 & 84.58 & 0.00 & 0.00 & 90.60 & 87.27 & -- \\
\midrule
\multirow{3}{*}{X351} & best & 26622.42 & 27064.88 & 27222.96 & 28008.09 & 26622.42 & 28639.35 & 28394.79 & 27204.23 & \textbf{26370.42} & -- \\
 & mean & 26750.60 & 27217.77 & 27498.03 & 28336.07 & 26750.60 & 28932.86 & 28571.17 & 27447.89 & \textbf{26448.67} & -1.13\% \\
 & std. & -- & 86.20 & 155.62 & 205.29 & 102.55 & 132.28 & 89.85 & 175.23 & 58.10 & -- \\
\midrule
\multirow{3}{*}{X459} & best & 24794.35 & 25370.80 & 25464.84 & 26048.21 & 24794.35 & 27371.06 & 27371.06 & 25615.60 & \textbf{24465.89} & -- \\
 & mean & 25041.10 & 25582.27 & 25809.47 & 26345.12 & 25041.10 & 27371.06 & 27371.06 & 25942.22 & \textbf{24600.78} & -1.76\% \\
 & std. & -- & 106.89 & 157.97 & 185.14 & 237.58 & 0.00 & 0.00 & 140.26 & 79.95 & -- \\
\midrule
\multirow{3}{*}{X573} & best & 51436.90 & 52181.51 & 51929.24 & 54189.62 & 51436.90 & 56884.09 & 57018.02 & 52458.56 & \textbf{51234.65} & -- \\
 & mean & 51776.70 & 52548.09 & 52793.66 & 55327.62 & 51776.70 & 57342.53 & 57417.98 & 52748.36 & \textbf{51379.40} & -0.77\% \\
 & std. & -- & 278.85 & 577.24 & 548.05 & 166.86 & 240.17 & 248.69 & 253.97 & 107.62 & -- \\
\midrule
\multirow{3}{*}{X685} & best & 69955.95 & 71345.40 & 72549.90 & 73925.56 & 69955.95 & 78045.96 & 78045.96 & 73975.35 & \textbf{69415.64} & -- \\
 & mean & 70401.25 & 71770.57 & 73124.98 & 74508.03 & 70401.25 & 78045.96 & 78045.96 & 74412.32 & \textbf{69610.37} & -1.12\% \\
 & std. & -- & 197.08 & 320.07 & 409.43 & 218.98 & 0.00 & 0.00 & 253.08 & 123.60 & -- \\
\midrule
\multirow{3}{*}{X749} & best & 79779.87 & 81002.01 & 81392.78 & 84034.73 & 79779.87 & 85781.59 & 85640.27 & 81939.07 & \textbf{78809.51} & -- \\
 & mean & 80135.67 & 81327.39 & 81848.13 & 84759.79 & 80135.67 & 85781.59 & 85767.46 & 82657.87 & \textbf{78937.48} & -1.50\% \\
 & std. & -- & 176.19 & 275.26 & 376.10 & 219.50 & 0.00 & 44.69 & 368.48 & 90.58 & -- \\
\midrule
\multirow{3}{*}{X819} & best & 161924.79 & 164289.95 & 165069.77 & 170965.68 & 161924.79 & 174375.10 & 173145.29 & 165577.34 & \textbf{159864.38} & -- \\
 & mean & 162530.67 & 164926.41 & 165895.78 & 172410.12 & 162530.67 & 174840.49 & 173596.85 & 166464.86 & \textbf{160001.36} & -1.56\% \\
 & std. & -- & 318.62 & 403.70 & 568.58 & 289.41 & 230.07 & 276.31 & 507.42 & 175.36 & -- \\
\midrule
\multirow{3}{*}{X916} & best & 336717.71 & 341649.91 & 342796.88 & 357391.57 & 336717.71 & 362477.43 & 360138.18 & 343838.22 & \textbf{332309.16} & -- \\
 & mean & 337641.92 & 342460.70 & 343533.85 & 360269.94 & 337641.92 & 363118.31 & 360745.35 & 345689.82 & \textbf{332558.20} & -1.51\%  \\
 & std. & -- & 510.66 & 556.98 & 1192.57 & 461.47 & 489.37 & 423.35 & 1147.38 & 174.34 & -- \\
\midrule
\multirow{3}{*}{X1001} & best & 75469.29 & 77476.36 & 78053.86 & 78832.90 & 75469.29 & 80523.17 & 80523.17 & 78075.73 & \textbf{74893.30} & -- \\
 & mean & 75931.28 & 77920.52 & 78593.50 & 79163.34 & 75931.28 & 80523.17 & 80523.17 & 78897.32 & \textbf{75130.68} & -1.05\%  \\
 & std. & -- & 234.73 & 306.27 & 229.19 & 304.10 & 0.00 & 0.00 & 345.21 & 135.43 & -- \\
\midrule
 w/t/l & vs b-LAHC &  & 2/2/13 & 1/2/14 & 1/2/14 & 4/2/11 & 1/2/14 & 1/2/14 & 3/2/12 & &  \\
\midrule
 rank &  &  & 3.53 & 4.59 & 5.59 & 2.59 & 7.00 & 6.59 & 3.94 & 2.18 & \\
\bottomrule
\end{tabular}
\end{table*}
\subsection{Benchmark Suite and Reference Methods}

We evaluate b-LAHC on the IEEE WCCI-2020 EVRP benchmark~\cite{mavrovouniotis2020techreport}, covering 17 instances (7 small E-set, 10 large X-set) under two commonly used termination criteria (Max Evals and Max Time). Table~\ref{tab:instance} summarizes the key characteristics of all instances. These instances are divided into two categories:
\begin{itemize}
    \item Each instance in E-set contains no more than 100 customers. For these instances, upper bounds (UB) are provided as baseline results, some of which are optimal.
    \item Each instance in X-set contains between 142 and 1,000 customers, and no known upper bounds are provided.
\end{itemize}

These instances present different characteristics in customers' spatial distribution and their demands, significantly affecting the structure of feasible solutions.


Two different computational budgets are available. The first one is defined as a multiple of the problem size $pz$, where $pz = |\{d\}| + |\mathcal{V}_c| + |\mathcal{V}_f|$. The maximum number of evaluations is given by
\begin{equation}
\text{Max Evals} = 25{,}000 \times pz
\notag
\label{eq:max-evals}
\end{equation}
The evaluation of a solution requires $\mathcal{O}(pz)$ time. In the official evaluator, each access to an arc weight $d_{ij}$ consumes a fraction $1/pz$ of the budget, meaning that even neighborhood exploration contributes to the overall evaluation count.

The second budget is defined in~\cite{jia2021bilevel, jia2022confidence} as:
\begin{equation}
\text{Max Time} = \omega \cdot \frac{|V_c| + |V_f|}{100} \quad \text{(hours)}
\notag
\label{eq:exectime}
\end{equation}
where the parameter $\omega$ is set to 1, 2, and 3 for instance groups E22–E101, X143–X916, and X1001, respectively.


Max Evals facilitates a fair comparison across algorithms. In contrast, Max Time is subject to differences in hardware, programming languages, compiler settings, and parallelization strategies. These factors make it difficult to quantify the computational effort and may lead to inconsistent performance assessments. Both criteria are adopted in this study. 

Table~\ref{tab:method-summary} summarizes the state-of-the-art methods for the benchmark. Among them, the VNS, SA, and GA implementations correspond to the top three algorithms submitted to the WCCI-2020 EVRP competition.

\begin{figure*}
  \centering
  \includegraphics[width=\textwidth]{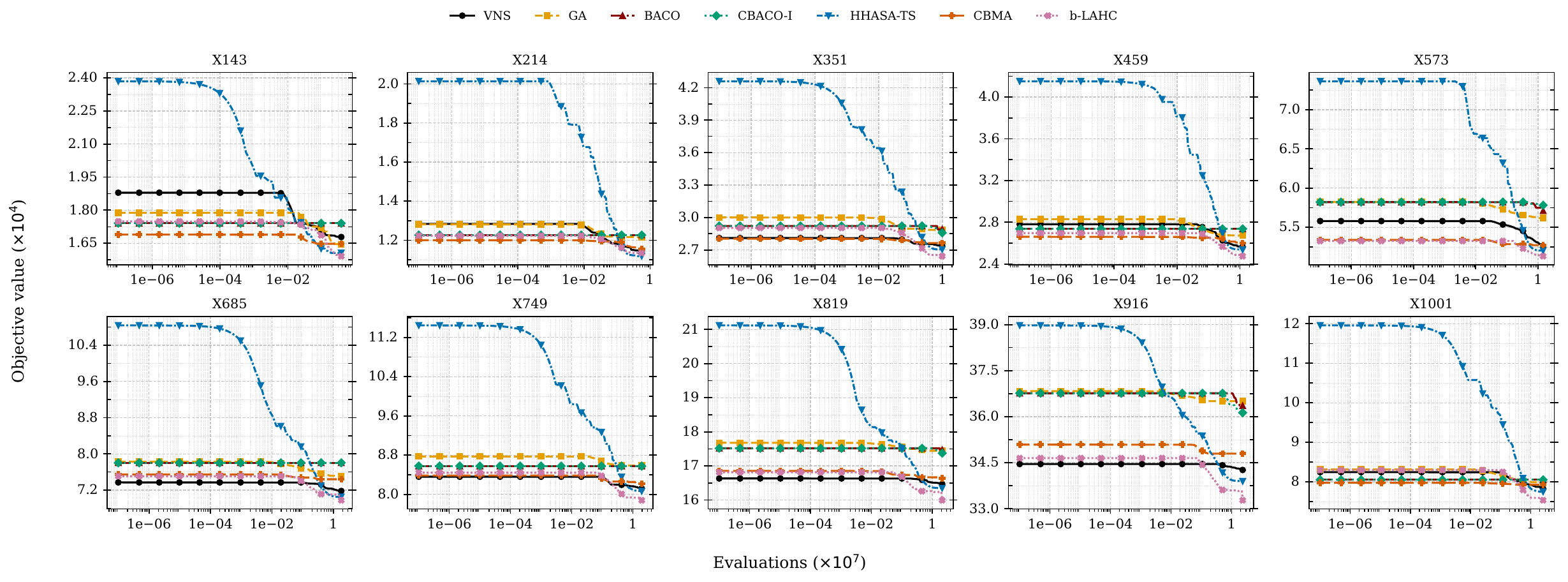}
  \caption{Convergence trends on the large-scale instances.}
  \label{fig:convergence-curves}
\end{figure*}

\subsection{Experimental Setup}

Table~\ref{tab:parameters} summarizes the control parameters used in b-LAHC. Default settings were obtained via automatic configuration using \textsc{irace}~\cite{lopez2016irace} on the full benchmark set. 


All experiments were conducted on the Sulis HPC platform using dedicated AMD EPYC 7742 nodes (2.25\,GHz, 64-core processors), restricted to single-thread execution and 1\,GB RAM per run. Unless otherwise specified, the budget is based on \emph{Max Evals} to ensure a fair comparison across algorithms. All available algorithms listed in Table~\ref{tab:method-summary} were evaluated under strictly identical experimental protocols, including hardware, programming language (C++), compiler (GCC 13.3.0), flag(\texttt{-O3}), and 10 independent runs per instance. The only exception was HHASA-TS, which was executed using its native MATLAB implementation due to unavailable source-level porting. For SA, results are taken directly from the official competition report, while TAMLS results are reproduced directly from its original paper.

For statistical significance testing, we employed the non-parametric Friedman test ($\alpha=0.05$), followed by a Holm post-hoc correction to account for multiple comparisons.

\subsection{Correlation vs. Misalignment: Validating $\phi(x)$ as a Surrogate Objective}

To empirically validate that the surrogate routing cost $\phi(x)$ is positively correlated with the complete objective $F(x, y^*(x))$, we collected unique $(\phi(x), F(x, y^*(x)))$ pairs by recording solutions encountered during the search process of b-LHAC. 

Table~\ref{tab:kendall-recall} summarizes the results in terms of Kendall’s $\tau_b$ rank correlation and Top-$k$\% set recall between $\phi(x)$ and $F(x, y^*(x))$. Across all 17 instances, the Kendall’s $\tau_b$ values are on average above 0.97, with $\tau_b > 0.95$ in 15 out of 17 instances. This confirms a very strong positive correlation between $\phi(x)$ and $F(x, y^*(x))$. 

However, the Top-$k$\% set recall reveals a more nuanced picture. While Recall@10\% and Recall@20\% remain above 0.95 in most instances, Recall@1\% exhibits noticeable drops in several large-scale X-instances (e.g., 0.0278 for X573 and 0.3169 for X819), even when $\tau_b$ is high (0.9811 and 0.9661, respectively). This discrepancy indicates a phenomenon of \emph{objective misalignment}. Although $\phi(x)$ and $F(x, y^*(x))$ are largely well-aligned, they may disagree on the precise ranking of top-performing solutions. In other words, solutions that appear optimal under $\phi(x)$ are not necessarily optimal under $F(x, y^*(x))$. Nevertheless, the optimal or near-optimal complete solutions under $F(x,y^*(x))$ are highly likely to be found within the broader top-$5\%$ or top-$10\%$ regions defined by $\phi(x)$.

In summary, using $\phi(x)$ as a cheap yet reliable surrogate objective during the search, significantly reduces lower-level evaluations while preserving solution quality. However, the observed \emph{objective misalignment} highlights the necessity of a unified bilevel optimization framework, as a naive two-stage approach may yield solutions far from the true optimal complete solution.

\subsection{Comparison with State-of-the-art Algorithms}

\subsubsection{Solution Quality under Max Evals}
Table~\ref{tab:max-evals} reports a head-to-head comparison of b-LAHC against state-of-the-art algorithms. ``BKS'' denotes the best-known solution. ``Gap'' is computed as the percentage difference between the b-LAHC mean and the BKS (negative means b-LAHC outperforms BKS). ``w/t/l'' counts the wins, ties, and losses of a compared algorithm against b-LAHC across all instances. ``rank'' shows the average rank over all instances (lower is better) based on the Friedman test.

\begin{itemize}[leftmargin=1em, itemsep=0pt]
    \item Small-scale (E-set, 7 instances). b-LAHC beats BKS on E33, ties BKS on E23 and E30, and is very close on the remaining four, with the largest deviation +0.71\% on E76. 
    \item Large-scale (X-set, 10 instances). b-LAHC surpasses BKS on 9 out of 10 instances and underperforms only on X214 (+0.35\%). These results confirm that b-LAHC scales particularly well with instance size.
    \item Statistical significance and overall ranking. A Friedman test over all 17 instances and 8 algorithms yields $\chi^2 = 74.67$ with $p = 1.67\times10^{-13} < 0.05$, rejecting the null hypothesis and indicating statistically significant performance differences. The overall ranking is: $\text{b-LAHC} \succ \text{HHASA-TS} \succ \text{VNS} \succ \text{CBMA} \succ \text{SA} \succ \text{GA} \succ \text{CBACO-I} \succ \text{BACO}$. 
    \item Head-to-head against the second-best method (HHASA-TS). b-LAHC achieves 11 wins / 2 ties / 4 losses at the mean level. On the winning subset, b-LAHC’s average gap to BKS is -0.99\%; on the losing subset it is +0.33\%. b-LAHC shows lower standard deviation than HHASA-TS on 12 out of 17 instances overall (3 higher, 2 ties), including 9 out of 10 large X-instances. Averaged over non-zero HHASA-TS cases, b-LAHC reduces standard deviation by 39.5\% overall (22.8\% on E-instances; 46.2\% on X-instances). This indicates that b-LAHC is not only more effective on most large X-instances but also more robust.
    \item Takeaway. Under the Max Evals budget, b-LAHC delivers near-BKS or better results on small E-instances and shows a clear advantage on large X-instances, achieving the top overall Friedman ranking with strong statistical significance.
\end{itemize}

\begin{table}[!t]
\centering
\caption{Ablation study on the effects of Greedy Descent ($\mathcal{G}$), Charging Refinement ($\mathcal{F}$), and Bilevel framework}
\label{tab:ablation}
\scriptsize
\setlength{\tabcolsep}{1pt}
\renewcommand{\arraystretch}{0.8}
\begin{tabular}{llccccccc}
\toprule
\multicolumn{1}{c}{\multirow{2}{*}{Instance}} &
\multicolumn{1}{c}{\multirow{2}{*}{Index}} &
\multicolumn{1}{c}{\multirow{2}{*}{Baseline}} &
\multicolumn{2}{c}{Remove $\mathcal{G}$} &
\multicolumn{2}{c}{Remove $\mathcal{F}$} &
\multicolumn{2}{c}{$\gamma = 0$} \\
\cmidrule(lr){4-5} \cmidrule(lr){6-7} \cmidrule(lr){8-9}
 &  &  & {\scriptsize Obj} & {\scriptsize Gap} & {\scriptsize Obj} & {\scriptsize Gap} & {\scriptsize Obj} & {\scriptsize Gap} \\
\midrule
\multirow{3}{*}{E22} & best & 384.67 & 384.67 & -- & 384.67 & -- & 385.39 & -- \\
 & mean & 385.25 & 384.96 & -0.08\% & 385.25 & 0.00\% & 394.81 & 2.48\% \\
 & std. & 0.28 & 0.35 & -- & 0.30 & -- & 13.61 & --  \\
\midrule
\multirow{3}{*}{E23} & best & 571.94 & 571.94 & -- & 571.94 & -- & 571.94 & -- \\
 & mean & 571.94 & 571.94 & 0.00\% & 571.94 & 0.00\% & 579.39 & 1.30\% \\
 & std. & 0.00 & 0.00 & -- & 0.00 & -- & 18.94 & -- \\
\midrule
\multirow{3}{*}{E30} & best & 509.47 & 509.47 & -- & 509.47 & -- & 509.47 & -- \\
 & mean & 509.47 & 511.26 & 0.35\% & 509.47 & 0.00\% & 530.06 & 4.04\% \\
 & std. & 0.00 & 1.98 & -- & 0.00 & -- & 20.49 & -- \\
\midrule
\multirow{3}{*}{E33} & best & 840.14 & 840.14 & -- & 840.14 & -- & 845.57 & -- \\
 & mean & 840.14 & 840.47 & 0.04\% & 840.14 & 0.00\% & 870.16 & 3.57\% \\
 & std. & 0.00 & 0.22 & -- & 0.00 & -- & 14.54 & -- \\
\midrule
\multirow{3}{*}{E51} & best & 529.90 & 535.44 & -- & 529.90 & -- & 529.90 & -- \\
 & mean & 531.19 & 538.47 & 1.37\% & 531.19 & 0.00\% & 568.05 & 6.94\% \\
 & std. & 3.59 & 2.47 & -- & 3.79 & -- & 19.51 & -- \\
\midrule
\multirow{3}{*}{E76} & best & 696.27 & 723.91 & -- & 696.27 & -- & 722.96 & -- \\
 & mean & 699.54 & 727.54 & 4.00\% & 699.54 & 0.00\% & 733.92 & 4.91\% \\
 & std. & 2.91 & 2.31 & -- & 3.07 & -- & 11.07 & -- \\
\midrule
\multirow{3}{*}{E101} & best & 840.55 & 880.66 & -- & 840.55 & -- & 865.80 & -- \\
 & mean & 844.01 & 882.26 & 4.53\% & 844.32 & 0.04\% & 894.58 & 5.99\% \\
 & std. & 3.16 & 1.07 & -- & 3.67 & -- & 16.10 & -- \\
\midrule
\multirow{3}{*}{X143} & best & 15916.19 & 16794.26 & -- & 15916.19 & -- & 17276.88 & -- \\
 & mean & 16103.45 & 16922.98 & 5.09\% & 16110.37 & 0.04\% & 17794.60 & 10.50\% \\
 & std. & 85.93 & 66.97 & -- & 81.94 & -- & 321.14 & -- \\
\midrule
\multirow{3}{*}{X214} & best & 11114.46 & 11202.04 & -- & 11136.01 & -- & 11823.46 & -- \\
 & mean & 11245.67 & 11307.11 & 0.55\% & 11252.84 & 0.06\% & 12104.47 & 7.64\% \\
 & std. & 87.27 & 52.07 & -- & 88.41 & -- & 133.49 & -- \\
\midrule
\multirow{3}{*}{X351} & best & 26370.42 & 26314.51 & -- & 26383.02 & -- & 28110.86 & -- \\
 & mean & 26448.67 & 26448.35 & 0.00\% & 26488.96 & 0.15\% & 28310.11 & 7.04\% \\
 & std. & 58.10 & 77.10 & -- & 56.99 & -- & 134.67 & -- \\
\midrule
\multirow{3}{*}{X459} & best & 24465.89 & 25510.22 & -- & 24486.95 & -- & 26309.21 & -- \\
 & mean & 24600.78 & 26566.49 & 7.99\% & 24612.92 & 0.05\% & 26697.30 & 8.52\% \\
 & std. & 79.95 & 492.94 & -- & 92.52 & -- & 192.83 & -- \\
\midrule
\multirow{3}{*}{X573} & best & 51234.65 & 67566.80 & -- & 51308.77 & -- & 52868.26 & -- \\
 & mean & 51379.40 & 73086.69 & 42.25\% & 51540.36 & 0.31\% & 53563.39 & 4.25\% \\
 & std. & 107.62 & 3506.93 & -- & 149.05 & -- & 395.22 & -- \\
\midrule
\multirow{3}{*}{X685} & best & 69415.64 & 85787.12 & -- & 70429.18 & -- & 73197.36 & -- \\
 & mean & 69610.37 & 94291.09 & 35.46\% & 70634.31 & 1.47\% & 73491.89 & 5.58\% \\
 & std. & 123.60 & 4974.22 & -- & 150.31 & -- & 193.54 & -- \\
\midrule
\multirow{3}{*}{X749} & best & 78809.51 & 93402.93 & -- & 79139.04 & -- & 82928.79 & -- \\
 & mean & 78937.48 & 97470.40 & 23.48\% & 79282.97 & 0.44\% & 83497.56 & 5.78\% \\
 & std. & 90.58 & 2183.00 & -- & 88.20 & -- & 308.47 & -- \\
\midrule
\multirow{3}{*}{X819} & best & 159864.38 & 160299.64 & -- & 161950.84 & -- & 164463.38 & -- \\
 & mean & 160001.36 & 161290.69 & 0.81\% & 162238.74 & 1.40\% & 164893.76 & 3.06\% \\
 & std. & 175.36 & 669.36 & -- & 166.57 & -- & 203.64 & -- \\
\midrule
\multirow{3}{*}{X916} & best & 332309.16 & 339313.98 & -- & 335162.52 & -- & 339887.86 & -- \\
 & mean & 332558.20 & 341834.68 & 2.79\% & 335479.71 & 0.88\% & 340409.34 & 2.36\% \\
 & std. & 174.34 & 1583.63 & -- & 261.13 & -- & 478.17 & -- \\
\midrule
\multirow{3}{*}{X1001} & best & 74893.30 & 326525.74 & -- & 75312.64 & -- & 80837.17 & -- \\
 & mean & 75130.68 & 360120.73 & 379.33\% & 75594.82 & 0.62\% & 82010.74 & 9.16\% \\
 & std. & 135.43 & 20775.16 & -- & 160.11 & -- & 497.33 & -- \\
\bottomrule
\end{tabular}
\end{table}

\subsubsection{Convergence Analysis}
Figure~\ref{fig:convergence-curves} shows convergence on large-scale X-instances with a log-scaled X-axis, which highlights early convergence. b-LAHC shows no advantage in the early phase: performing worse than CBMA (X143, X214, X459), VNS (X685--X916), and CBMA, CBACO-I, and BACO (X1001). This suggests that these algorithms benefit from stronger initialization, whereas b-LAHC starts from a randomly generated solution and relies on the greedy descent ($\mathcal{G}$) phase to reach a local optimum. In the later phase, however, b-LAHC clearly dominates on all instances except for X214. This strongly indicates that the success of b-LAHC lies in the strength of its bilevel framework design, despite its weaker initialization.

\subsubsection{Solution Quality under Max Time}

Under the Max Time setting, a Friedman test conducted across all $17$ instances and $5$ algorithms yields $\chi^2 = 40.78$ with a $p$-value of $2.99\times10^{-8} < 0.05$, indicating statistically significant performance differences. The average ranking is $\text{b-LAHC} \succ \text{TAMLS} \succ \text{CBMA} \succ \text{CBACO-I} \succ \text{BACO}$, where b-LAHC achieves the lowest average rank and consistently outperforms competing methods, particularly on large X-instances. Detailed experimental results are provided in the Supplementary Material.

\begin{figure}[t]
  \centering
  \includegraphics[width=1.0\linewidth]{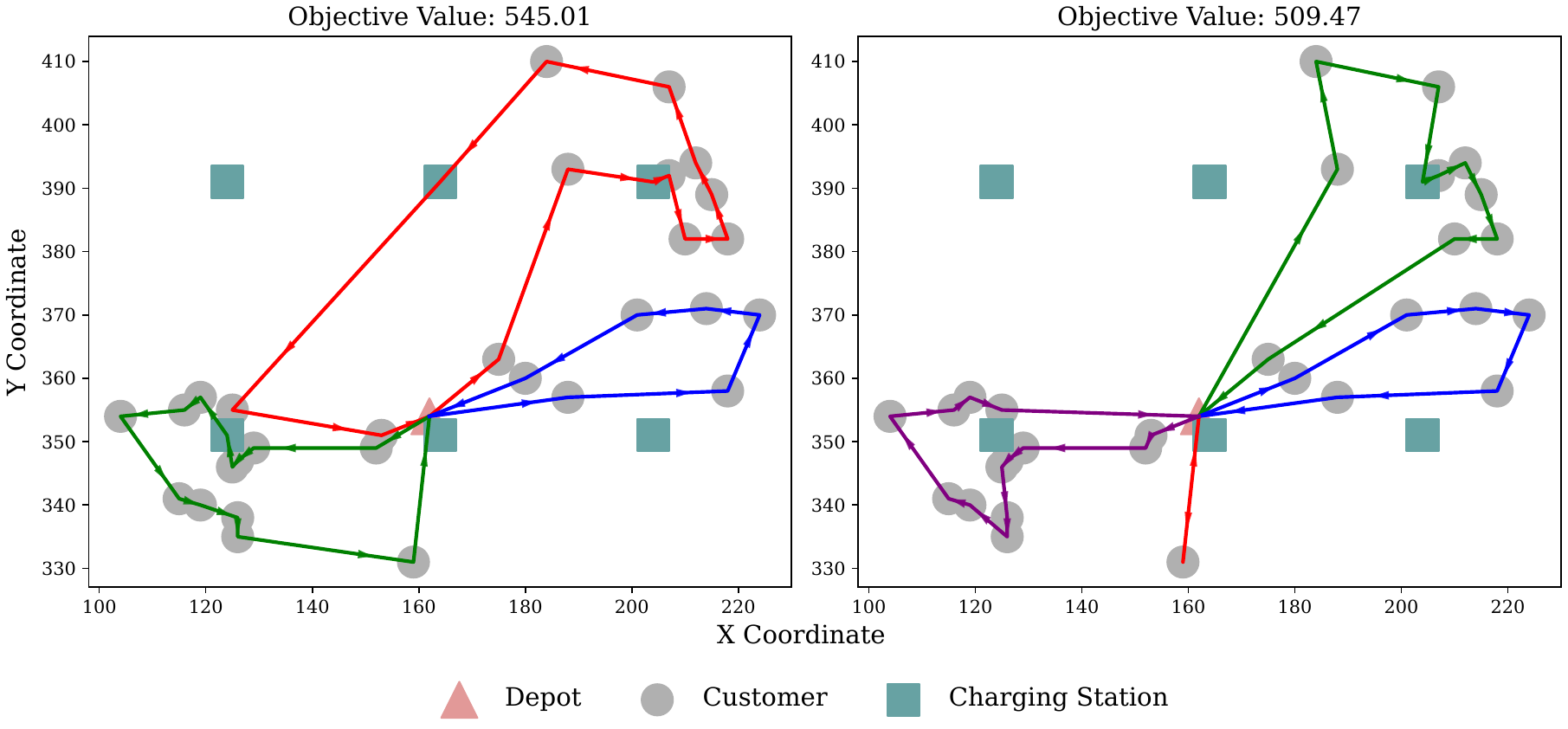}
  \caption{Comparative solution with and without M8 operator on E30.}
  \label{fig:M8-operator}
\end{figure}

\subsection{Ablation Study on Key Components}

Table~\ref{tab:ablation} summarizes the results of the ablation study, evaluating the contributions of the Greedy Descent ($\mathcal{G}$) and Charging Refinement ($\mathcal{F}$) components. In addition, we also examine the case $\gamma=0$, where b-LAHC degenerates into a naive two-stage pipeline with upper- and lower-level decisions solved separately. The baseline corresponds to b-LAHC executed with the parameters settings given in Table~\ref{tab:parameters}.

Removing $\mathcal{G}$ causes a substantial degradation in performance on 15 out of 17 instances, with an average mean gap of $+29.88\%$ relative to the baseline, and a median gap of $+2.79\%$. The deterioration is especially severe on large-scale instances such as X1001 and X573, while the impact on small E-instances is negligible.  

Removing $\mathcal{F}$ has a moderate effect but still worsens performance on 13 out of 17 instances, with an average mean gap of $+0.32\%$ and a median gap of only $+0.05\%$. The influence of $\mathcal{F}$ is negligible on small E-instances but becomes critical on large-scale cases where charging refinement is needed. For instance, excluding $\mathcal{F}$ results in noticeable degradations on X685, X819, and X916. It is also worth noting that incorporating $\mathcal{F}$ incurs only a modest computational overhead. Enabling $\mathcal{F}$ adds less than $0.1$\,s of runtime for all instances. This negligible overhead justifies the inclusion of $\mathcal{F}$ given its significant improvement in solution quality.

Setting $\gamma=0$ leads to a consistent deterioration across nearly all instances, highlighting the importance of the bilevel framework. The average mean gap increases to $+5.48\%$ with a median gap of $+5.58\%$, markedly higher than the baseline. The degradation is already visible on small E-instances, and it becomes even more pronounced on large-scale cases such as X143, X351, and X1001. Compared with removing $\mathcal{F}$, the impact of disabling the bilevel framework is significantly larger, underscoring the necessity of jointly optimizing upper- and lower-level decisions rather than treating them in a decoupled two-stage manner.

Overall, these findings indicate that $\mathcal{G}$ and $\mathcal{F}$ play complementary roles, while the bilevel framework itself is indispensable. Their synergy enables the full b-LAHC algorithm to achieve superior solution quality, scalability, and stability across diverse problem scales.

The neighborhood exploration initially comprised M1--M7, all of which preserve or reduce the number of routes. They led to unstable behavior in certain instances. In particular, on E30 (Fig.~\ref{fig:M8-operator}), some runs converged to inferior three-route solutions (objective value $545.01$), whereas the best-known outcomes are four-route solutions ($509.47$). This counter-intuitive phenomenon arises from the EV setting, where charging detours and route geometry can favor additional routes.

To address this, we introduced M8, the only move able to \emph{increase} the route count. M8 explores regions of the solution space that are unreachable by M1--M7 and provides an escape from local optima caused by under-utilized fleets. Empirically, enabling M8 stabilizes convergence on E30 (Fig.~\ref{fig:M8-operator}), with runs consistently reaching the superior four-route solution, and we observed analogous improvements in other cases.

\subsection{Sensitivity Analysis of Hyperparameters}
\label{subsec:sensitivity}


The optimal choice of $L_h$ depends strongly on problem scale, consistent with Lobo~\etal~\cite{lobo2020}. Very small values lead to premature convergence, whereas excessively large values slow progress under a fixed budget. A moderate setting provides the best trade-off, and $L_h=5723$ is adopted as it yields stable performance across instances. $\eta_{\max}$ controls neighborhood exploration effort. The algorithm shows strong robustness to this parameter; overly small values reduce exploration on large instances, while overly large values waste evaluations on small ones. The default setting $\eta_{\max}=60$ achieves a balanced performance. $\gamma$ regulates how frequently the lower-level optimizer is invoked. Experiments indicate stable performance when $\gamma \in [1.01,\,1.05]$, which balances upper-level and lower-level decisions. We therefore use $\gamma=1.01$ as the default setting. Detailed sensitivity analyses for these parameters are provided in the Supplementary Material.

\section{Conclusions and Future Work}
\label{sec:conclusion}

This paper introduced a bilevel late acceptance hill climbing algorithm (b-LAHC) for the bilevel formulation of the E-CVRP. By leveraging a surrogate objective at the upper level with an efficient charging optimizer at the lower level, b-LAHC is lightweight and interpretable. Despite using fixed parameters, it delivers efficient search across instances.

Compared with state-of-the-art methods, b-LAHC follows a distinct design philosophy. HHASA-TS~\cite{rodriguez2024new} integrates SA with reinforcement learning to dynamically adjust the search, improving adaptiveness but incurring computational overhead and higher variability on large instances. CBMA~\cite{qin2024confidence} and CBACO-I~\cite{jia2022confidence} employ adaptive diversity control in population-based frameworks, enabling effective exploration on small/medium instances but showing slower convergence and reduced robustness on larger benchmarks. Notably, b-LAHC attains comparable or superior results as a lightweight single-point method without relying on complex adaptive mechanisms. These findings suggest two directions: incorporating adaptive learning into b-LAHC or incorporating promising features in b-LAHC into a population-based framework.

Interestingly, M8 operator plays a more critical role than expected. In some instances (e.g., E30), optimal solutions require more vehicles due to charging detours and battery limits. Without M8, the search converges prematurely to suboptimal solutions featuring fewer vehicles. This highlights the importance of route-creation moves for handling complex EV charging behaviors.

The experiments provide strong support for the proposed bilevel framework, which leverages the dominance of upper-level routing decisions in the E-CVRP, while offering a principled mechanism to determine when to introduce joint optimization, thereby ensuring greater generality and adaptability across diverse problem instances.

Future research can proceed in three main directions. First, we plan to extend the current b-LAHC algorithm by incorporating learning mechanisms to enable adaptive behavior during the search. Second, we aim to incorporate useful features in the single-point framework into a population-based metaheuristic capable of multimodal search, allowing a systematic comparison between single-point and population-based strategies. Third, the proposed bilevel optimization model can be applied to other routing problems with hierarchical decision structures, offering a unified perspective for tackling bilevel routing challenges beyond the E-CVRP.
    
\section*{Acknowledgments}
The authors thank the IT support team at QMUL. Computations were performed on the Sulis Tier-2 HPC platform, funded by EPSRC (EP/T022108/1) and HPC Midlands+.  

\bibliographystyle{IEEEtran}
\bibliography{bibtex/bib/IEEEexample}

\end{document}